\documentclass[11pt]{article}

\usepackage[a4paper,margin=1in]{geometry}
\usepackage{graphicx}

\usepackage{cite}

\usepackage{subcaption}
\usepackage{float}
\usepackage{latexsym}
\usepackage{amssymb}
\usepackage{amsmath}
\usepackage{multirow}
\usepackage{adjustbox}	
\usepackage{tabularx}
\usepackage{bold-extra}
\usepackage{xcolor}
\usepackage{enumitem}
\usepackage{algorithm}
\usepackage[noend]{algpseudocode}
\usepackage{amsfonts}
\usepackage{array}

\usepackage{booktabs,arydshln}
\usepackage{hyperref}
\hypersetup{
    colorlinks=true,
    linkcolor=blue,
    filecolor=magenta,      
    urlcolor=cyan,
    pdfpagemode=FullScreen,
    }
    
\usepackage{cleveref}

\usepackage{acronym}

\acrodef{FSL}{Few-Shot Learning}
\acrodef{SSL}{Self-Supervised Learning}
\acrodef{DA}{Domain Adaptation}
\acrodef{MAML}{Model-Agnostic Meta-Learning}

\acrodef{OOD}{Out-Of-Domain}
\acrodef{ID}{In-Domain}

\acrodef{IDA}{ID adaptation}

\acrodef{UIAug}{Unsupervised Instance-based Augmentation}
\acrodef{UCKM}{Unsupervised Constrained K-Means}

\newcommand{\CAPITAN}{\textsc{Capitan}}
\newcommand{\Greek}{\textsc{Greek}}
\newcommand{\TKH}{\textsc{TKH}}
\newcommand{\Omniglot}{\textsc{Omniglot}}
\newcommand{\Egyptian}{\textsc{Egyptian}}
\newcommand{\OrganAMNIST}{\textsc{OrganAMNIST}}
\newcommand{\CIFAR}{\textsc{CIFAR-FS}}
\newcommand{\MiniImageNet}{\textsc{MiniImageNet}}

\DeclareMathOperator*{\argmax}{\arg\!\max}

\begin{document}

\title{Are We Really Doing Few-Shot Learning?\\A Critical Examination of Pre-Training Assumptions}


%
%


\author{
    Alejandro Galan-Cuenca \and
    Marcelo Saval-Calvo \and
    Antonio Javier Gallego\\[0.5em]
    \small University Institute for Computer Research, University of Alicante, Alicante, Spain\\
    \small \texttt{a.galan@ua.es, m.saval@ua.es, jgallego@dlsi.ua.es}
}

\date{}

\maketitle

\begin{abstract}
Few-shot learning is commonly evaluated under protocols that pre-train a model on a large auxiliary set whose classes are disjoint from the target episodes yet drawn from the same visual domain. This paper examines whether such protocols truly reflect low-data learning. We systematically compare no pre-training, class-disjoint in-domain pre-training, supervised out-of-domain pre-training, and label-free out-of-domain pre-training across eight datasets, three few-shot architectures, and multiple way-shot settings. Our results show that class disjointness alone is insufficient to remove the influence of target-domain data. In-domain pre-training improves over no pre-training by 33.41 percentage points on average, whereas supervised out-of-domain pre-training yields 23.75 percentage points, revealing a 9.66-point optimistic bias associated with domain overlap. Although out-of-domain pre-training is more realistic in applications where target-domain data are scarce, its effectiveness depends strongly on the compatibility between source and target domains. We further show that labeled source data are not strictly required, with an augmentation-based label-free strategy reaching an average gain of 27.71 percentage points and closely matching supervised out-of-domain pre-training at 27.97 percentage points. Finally, we introduce a descriptor-based source-selection strategy that estimates source-domain suitability before pre-training, reaching a median gap of only 1.37 percentage points to oracle selection. These findings highlight the need to move beyond in-domain pre-training as the default few-shot evaluation protocol, since it can overestimate performance in realistic scenarios where target-domain data are scarce.
\end{abstract}



\section{Introduction}
\label{sec:introduction}

\ac{FSL} is a subfield of machine learning focused on developing models that can learn and generalize from extremely limited training data~\cite{Wang:ACM:2020, Li:TPAMI:2006, Fink:neurips:2004}. In classification, it is commonly formalized through the $n$-way $k$-shot framework, where a model must discriminate among $n$ classes using only $k$ training examples per class. Standard settings include 5-way 1-shot and 20-way 5-shot, requiring classification among 5 or 20 classes using 1 or 5 samples per class, respectively. In this framework, data is organized into a \textit{support set}, containing the only labeled examples available as prior knowledge (i.e., the $n \times k$ training samples), and a \textit{query set}, containing unseen examples used to evaluate generalization. This \ac{FSL} setting naturally arises when collecting large volumes of labeled data is costly, limited, or impractical, including rare-species classification~\cite{Dasgupta:CVPR:2024, Zhang:EcologicalInformatics:2023, Villon:EcologicalInformatics:2021}, medical imaging and diagnosis with scarce annotated or pathological cases~\cite{Tang:ICCV:2021, Feng:IEEE:2021, Moor:pmlr:2023, Galan:PAA:2024}, fraud detection with few positive cases~\cite{Lai:KnowledgeBasedSystems:2025}, satellite-based Earth observation~\cite{Mohammadi:RemoteSensing:2024}, wideband signal detection and recognition~\cite{Hao:TWC:2024}, and facial recognition with few images per individual~\cite{Shome:ICCV:2021, Chen:IS:2023, Zhu:ESwA:2022}, among many other applications.

In the \ac{FSL} literature, it is common to assume, beyond the support and query sets, the availability of a labeled dataset for model pre-training. Standard benchmarks typically define class-disjoint splits: a test split containing the target classes, from which the support and query sets---commonly referred to as an \textit{episode} or \textit{meta-test set}---are constructed, and a pre-training split, commonly called the \textit{meta-train set}, containing the remaining classes and samples\footnote{Some benchmarks also define an additional validation split derived from the meta-training set, with disjoint classes, for model selection and hyperparameter tuning.}~\cite{Parnami:arxiv:2022, Song:ACM:2023}. Although this pre-training\footnote{We generally refer to the meta-training split as the pre-training set to emphasize its role in learning the feature extractor on a source dataset before episodic evaluation on novel classes.} set contains classes different from those in the episode, its samples come from the same domain as the target classes, resulting in an \ac{ID} class-disjoint pre-training scheme. For example, if the target classes are specific bird species, pre-training may use other bird species from the same dataset.

\newlength{\subfigH}
\setlength{\subfigH}{0.18\textheight} 
\begin{figure*}[!ht]
    \centering
    \begin{subfigure}[t]{.19\textwidth}
        \centering
        \includegraphics[height=\subfigH,keepaspectratio]{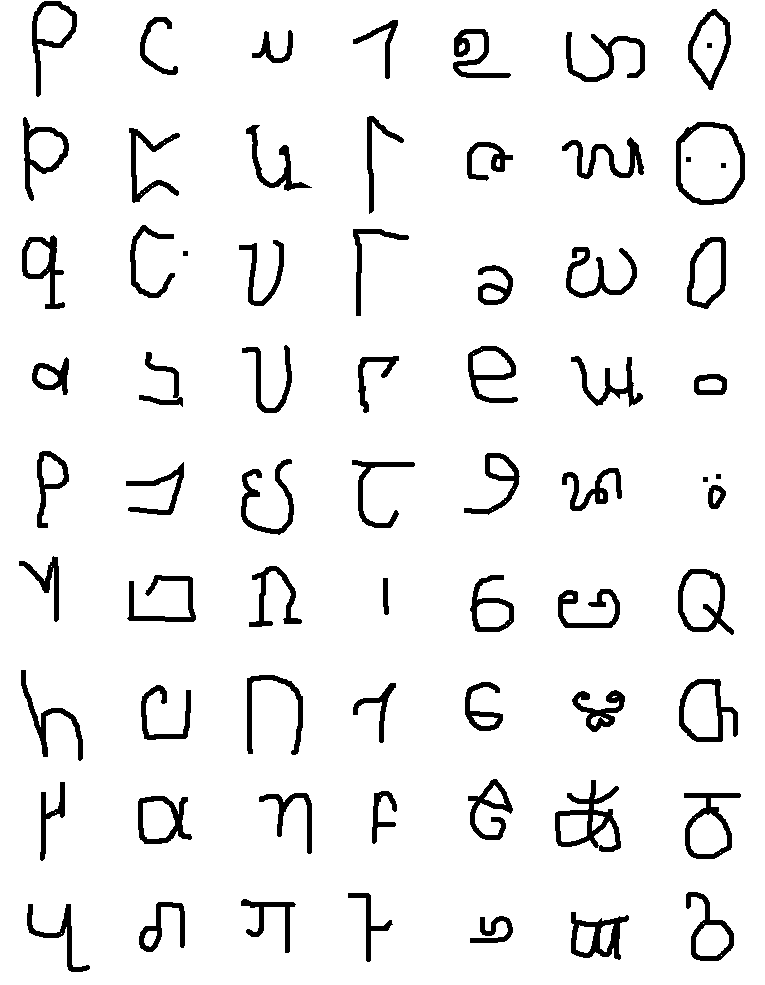}
        \caption{Omniglot examples.}
        \label{fig:intro_example:omniglot}
    \end{subfigure}\hfill
    \begin{subfigure}[t]{.38\textwidth}
        \centering
        \includegraphics[height=\subfigH,keepaspectratio]{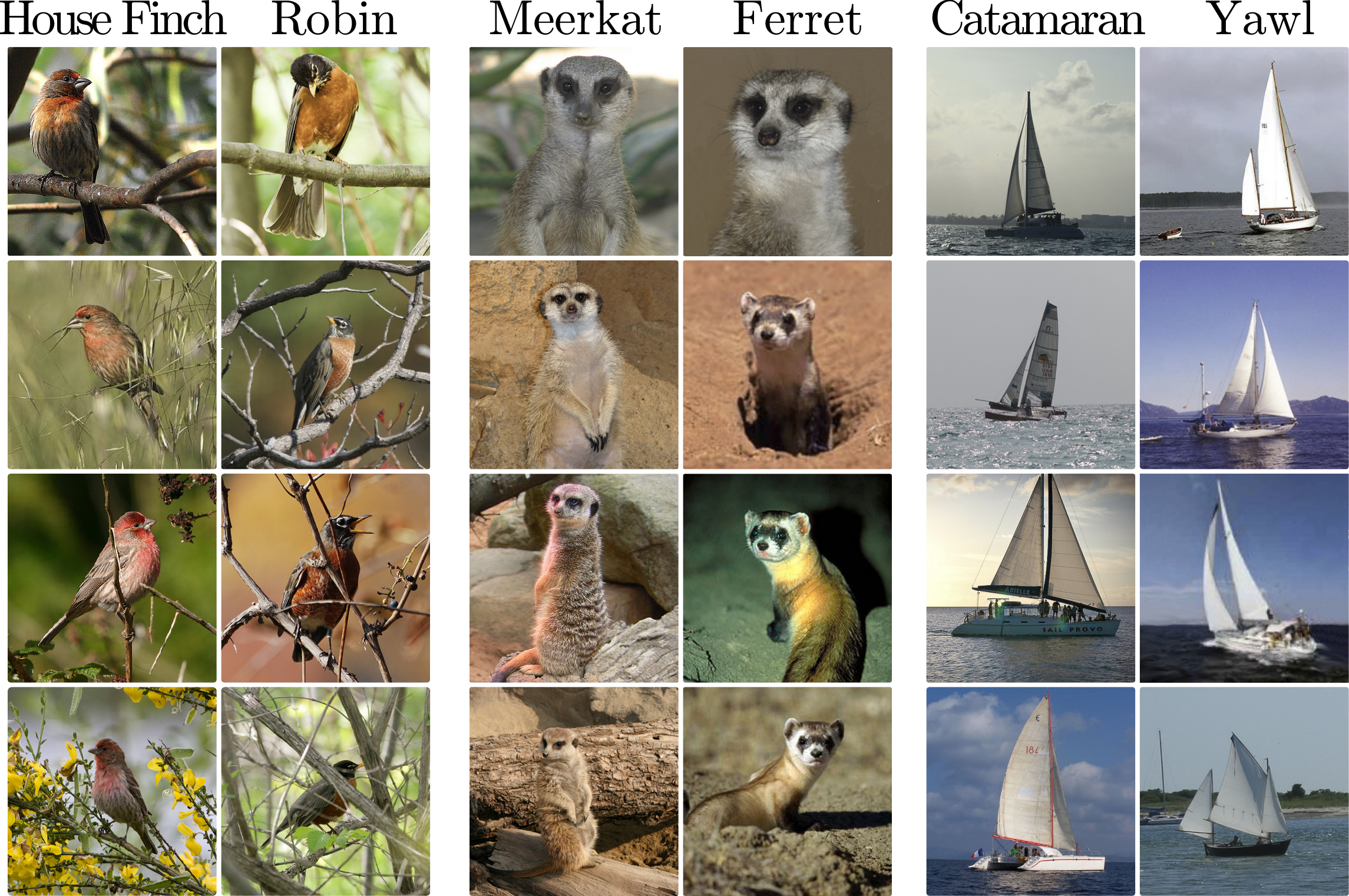}
        \caption{miniImageNet examples.}
        \label{fig:intro_example:imagenet}
    \end{subfigure}\hfill
    \begin{subfigure}[t]{.35\textwidth}
        \centering
        \includegraphics[height=\subfigH,keepaspectratio]{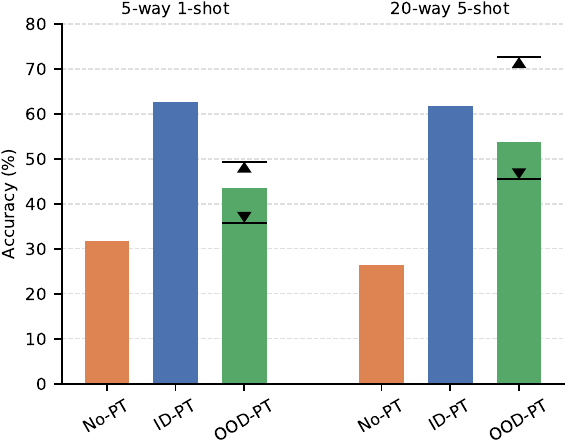}
        \caption{Omniglot 5/20-way 1/5-shot results.}
        \label{fig:intro_example:results}
    \end{subfigure}
    
    \caption{\ac{ID} similarity in standard \ac{FSL} benchmarks and its effect on pre-training. 
    (a) Omniglot: 63 characters spanning 50 alphabets, grouped by similarity; each displayed character corresponds to a distinct class.  
    (b) miniImageNet: examples of visually similar, semantically related classes. 
    (c) Comparison of pre-training (PT) strategies on Omniglot under 5-way 1-shot and 20-way 5-shot configurations: No-PT, ID-PT, and OOD-PT. For OOD-PT, bars report the mean accuracy over seven source domains, and arrows mark the best and worst source (max/min) accuracy.
    }    
    \label{fig:intro_example}
\end{figure*}

Although this scenario may appear artificial and idealized, it is the standard evaluation protocol in \ac{FSL} benchmarks, where datasets such as miniImageNet~\cite{Vinyals:NeurIPS:2016}, Omniglot~\cite{Lake:AAAS:2015}, and CIFAR-FS~\cite{Bertinetto:arxiv:2019} are partitioned following this methodology. Despite class-disjoint splits, the sets retain highly similar visual characteristics. Omniglot, for instance, contains 1,623 handwritten characters from 50 alphabets, yet all samples share similar black-and-white handwritten features (see Fig.~\ref{fig:intro_example:omniglot}). Similarly, miniImageNet includes color photographs from 100 classes with substantial visual overlap in shapes, textures, predominant colors, and backgrounds. Its classes also belong to semantically related groups, including vehicles, animals, plants, and objects, with closely related pairs such as house finch vs. robin, meerkat vs. ferret, and catamaran vs. yawl (see Fig.~\ref{fig:intro_example:imagenet}).

This \ac{ID} pre-training setup relaxes the core premise of \ac{FSL}, which is to learn when only a few labeled samples are available in the target domain. Although pre-training and target classes are disjoint, the model still benefits from the high visual similarity across classes, thereby reducing the task complexity. This raises two critical issues:

\begin{itemize}
    \item It introduces an optimistic bias by exploiting shared distributional properties between pre-training and target classes, thus resembling conventional supervised learning, where training and test data come from the same i.i.d. distribution.

    \item It is unrealistic to assume that in real-world \ac{FSL} applications, there will be abundant \ac{ID} labeled data from other (disjoint) classes, already annotated and sufficient to perform such a pre-training step. 
\end{itemize}

In practice, the pre-training step will involve non-i.i.d. conditions---i.e., \ac{OOD} data---since obtaining large amounts of labeled \ac{ID} data is difficult or even infeasible in some domains. Consequently, the pre-training distribution may differ substantially from that of the target episodes.

To assess this more realistic scenario, we compare three pre-training strategies: (i) \ac{ID} pre-training, (ii) \ac{OOD} pre-training, and (iii) no pre-training, where the model relies solely on the limited labeled samples in the episode. Our goal is to show that current \ac{FSL} benchmark protocols introduce a systematic bias, yielding overly optimistic and therefore unfair evaluations, and to establish a more robust and realistic framework for \ac{FSL} training and evaluation.

Fig.~\ref{fig:intro_example:results} illustrates the performance of the three pre-training (PT) strategies on Omniglot under 5-way 1-shot and 20-way 5-shot settings. The results show both the substantial benefit of pre-training (No-PT vs. ID/OOD-PT) and the optimistic advantage of ID-PT over OOD-PT. Note that OOD-PT results are averaged across seven heterogeneous source domains, ranging from similar handwritten character datasets to more distant natural and medical images (see Sect.~\ref{sec:setup}). In the 1-shot setting, ID-PT achieves 62.3\% accuracy, compared with a 43.4\% OOD-PT average and drops of up to 27 points depending on the source domain. In the 20-way 5-shot setting, ID-PT reaches 61.7\%, versus a 53.8\% OOD-PT average, with substantial variability across pre-training domains (45.5\%--72.6\%).

Beyond analyzing the optimistic bias of standard \ac{ID} pre-training, this work explores several strategies to cope with this issue: (i) a simple \ac{IDA} step---episodic fine-tuning on the scarce $n \times k$ labeled \ac{ID} support samples---to improve performance after \ac{OOD} pre-training; (ii) label-free pre-training based on unsupervised and \ac{SSL} techniques, eliminating the need for labeled pre-training data; and (iii) domain-similarity measures to rank candidate \ac{OOD} sources for a target and select, before pre-training, the one most likely to improve downstream few-shot performance.

Experiments were conducted using three widely adopted \ac{FSL} architectures---Matching Networks~\cite{Vinyals:NeurIPS:2016}, Prototypical Networks~\cite{Snell:NeurIPS:2017}, and Relation Networks~\cite{Sung:CVPR:2018}---which represent the field's main methodological approaches and have served as the foundation for many subsequent works. 
We evaluate all source–target combinations across eight datasets: three standard benchmarks---miniImageNet, Omniglot, and \CIFAR~\cite{Vinyals:NeurIPS:2016, Lake:AAAS:2015, Bertinetto:arxiv:2019}---and five additional datasets spanning Greek, Egyptian, and Korean text, musical scores, and medical images.

The remainder of this paper is organized as follows. Section~\ref{sec:related-work} reviews related work on \ac{FSL} and pre-training; Section~\ref{sec:method} presents the methodology and proposed approaches; Section~\ref{sec:setup} describes the datasets, evaluation protocols, and implementation details; and Section~\ref{sec:results} reports and discusses the results. Finally, Section~\ref{sec:conclusions} summarizes the main findings and future research directions.

\section{Related Work}
\label{sec:related-work}

Within the broad landscape of machine learning, \ac{FSL} has attracted considerable attention as a data-efficient learning paradigm for generalizing from few labeled examples. Comprehensive reviews of \ac{FSL} and related meta-learning approaches are provided in~\cite{Wang:ACM:2020, Song:ACM:2023, Vettoruzzo:TPAMI:2024, Gharoun:ACN:2024}.

Several families of methods have been proposed for few-shot generalization, including (i) data augmentation~\cite{Zhang:TNNLS:2025}, (ii) generative approaches that synthesize additional training examples~\cite{Luo:WACV:2021, Li:CVPR:2020}, (iii) transfer learning and \ac{DA} methods that adapt models to new target domains using labeled data from related sources or tasks~\cite{Zhao:WACV:2021}, and (iv) meta-learning, including optimization-based methods such as \ac{MAML}~\cite{Chelsea:ICML:2017} and metric-based approaches~\cite{Vinyals:NeurIPS:2016, Snell:NeurIPS:2017, Sung:CVPR:2018}. Among them, meta-learning offers a more flexible and powerful framework by learning inductive biases and similarity structures that enable discrimination from few labeled examples, rather than relying mainly on the appearance or characteristics of previously available data, as is typically the case in \ac{DA}, augmentation, and generative schemes. This section focuses on meta-learning-based \ac{FSL} methods, which are the primary target of this work.

In visual recognition, meta-learning has become a central framework for \ac{FSL}~\cite{Vettoruzzo:TPAMI:2024}. Rather than learning a single classifier over a fixed label set, it exploits experience across many tasks---typically subsets of classes---to build models that can be quickly adapted or are inherently effective on \textit{novel} tasks with few labeled examples. In metric-based methods, query labels are inferred by comparing query embeddings with the support set through attention mechanisms (Matching Networks~\cite{Vinyals:NeurIPS:2016}), class prototypes (Prototypical Networks~\cite{Snell:NeurIPS:2017}), or learned similarity functions (Relation Networks~\cite{Sung:CVPR:2018}). Numerous refinements and alternative baselines have since been proposed, including non-episodic and nearest-neighbor pipelines~\cite{Tian:ECCV:2020, Wang:arxiv:2019, Dhillon:ICLR:2020}, embedding adaptation methods~\cite{Ye:CVPR:2020}, and analyses of training and evaluation choices~\cite{Chen:arxiv:2020, Luo:ICML:2023}.

A key component of most meta-learning methods is \emph{episodic training}. Introduced by Vinyals et al.~\cite{Vinyals:NeurIPS:2016}, it mimics few-shot evaluation by sampling episodes from a larger training pool, each comprising an $n$-way $k$-shot support set and a query set\footnote{During meta-training, the query set contains additional \textit{labeled} examples from the same $n$ classes, whose labels are used to compute the episode loss; at meta-test time, query labels are used only for evaluation.}. Episodic training is now widely used in metric-based, optimization-based~\cite{Chelsea:ICML:2017, Lin:ESwA:2023}, and memory-augmented models~\cite{Santoro:ICML:2016, Mitchell:ICML:2022}.

Although episodic training has become closely associated with \ac{FSL}, recent work has questioned some of its assumptions. Laenen et al.~\cite{Laenen:NeurIPS:2021} show that episode construction and permutations affect training dynamics and that episodic sampling is not universally beneficial. Dhillon et al.~\cite{Dhillon:ICLR:2020} and Alshamrani et al.~\cite{Alshamrani:SPA:2023} further show that standard supervised training without episodic sampling can match or outperform episodic meta-learning in some settings, particularly as the number of training classes increases.

Beyond episode sampling, growing attention has focused on the \emph{data regime} used to develop and benchmark meta-learning methods. Standard benchmarks such as miniImageNet, Omniglot, and CIFAR-FS~\cite{Vinyals:NeurIPS:2016, Lake:AAAS:2015, Bertinetto:arxiv:2019} assume a large labeled dataset divided into class-disjoint meta-training, validation, and meta-testing splits. Although their classes do not overlap, all splits come from the same dataset and therefore share similar visual characteristics and semantics, yielding an \ac{ID} class-disjoint pre-training setup. Consequently, models are pre-trained on abundant labeled \ac{ID} data before few-shot evaluation, reducing task difficulty and introducing optimistic bias in reported performance (Fig.~\ref{fig:intro_example:results}). Thus, although \ac{FSL} targets low-data learning, much of the literature still assumes access to a relatively large meta-training set at least weakly aligned with the target tasks~\cite{Vinyals:NeurIPS:2016, Snell:NeurIPS:2017, Sung:CVPR:2018, Wang:arxiv:2019, Ye:CVPR:2020, Chen:arxiv:2020, Liu:ECCV:2020, Ziko:ICML:2020, Boudiaf:NeurIPS:2020, Dhillon:ICLR:2020, Tian:ECCV:2020, Hu:ICANN:2021, Lee:IEEE:2021, Luo:ICML:2023, Lin:ESwA:2023, Chen:Nature:2024, Luo:RESS:2024}, among many others.

To address this issue, several works move beyond the \ac{ID} regime by studying distribution shifts between meta-training and meta-testing data. This setting, commonly termed Cross-Domain \ac{FSL} (CD-FSL), assumes source and target tasks from different domains and, in the canonical formulation, disjoint label spaces. In practice, CD-FSL often trains on one dataset and evaluates on tasks from another, a protocol also known as Cross-Dataset \ac{FSL}~\cite{Xu:ACM:2025, Guo:ECCV:2020, Li:IEEE:2022, Li:arxiv:2021, Wang:IJCAI:2021, Cheng:ICCV:2023}. Most works focus on improving transfer through domain-invariant representations~\cite{Fu:ACMM:2021}, autoencoder-based disentanglement of domain-specific and invariant factors~\cite{Liang:ICCV:2021}, cross-domain distance functions~\cite{Li:IEEE:2022}, multi-source training~\cite{Li:IJCV:2024}, and pixel-level affinity mechanisms~\cite{Chen:CVPR:2024}.

However, using different datasets does not guarantee strict class disjointness or a genuinely \ac{OOD} setting. Several popular few-shot benchmarks share parent collections---for example, miniImageNet and tieredImageNet are both derived from ImageNet---yet are sometimes treated as distinct sources in cross-dataset protocols~\cite{Guo:ECCV:2020, Xu:ACM:2025}. Even with disjoint label sets, datasets may contain semantically overlapping categories; moreover, when both benchmarks derive from the same data source, the resulting shift may remain largely \ac{ID} in terms of visual statistics and semantics. To prevent label leakage and near-\ac{ID} shifts, we use completely disjoint datasets with explicitly non-overlapping class sets between pre-training and target tasks. This protocol also allows us to systematically analyze how the nature and degree of source--target domain similarity affect few-shot performance.

While CD-FSL methods typically assume a fixed or small set of source--target benchmarks and propose architectural or training modifications to improve cross-domain adaptation, we take a different perspective: when suitable \ac{ID} data are unavailable, cross-domain pre-training is often the default. The key questions are then \emph{which} \ac{OOD} source to use for a target and how to \emph{eliminate source-label requirements} so that any dataset can serve as pre-training data. Rather than proposing a new cross-domain meta-learner, we study the \emph{pre-training regime} itself by comparing \ac{ID}, truly \ac{OOD}, and no pre-training, and analyzing the effects of (i) source-dataset choice, (ii) source-label availability, and (iii) \ac{IDA}.

Recent works have also explored reducing reliance on labeled source data through \ac{SSL} pre-training before few-shot adaptation. Alfaro-Contreras et al.~\cite{AlfaroContreras:PRL:2023} study this setting for few-shot symbol classification, while Hu et al.~\cite{Hu:IEEE:2022} show that external data and simple fine-tuning can substantially influence few-shot performance. Our study complements this line by separating \emph{label supervision} from \emph{domain alignment}. We rank candidate \ac{OOD} sources using domain-similarity criteria before pre-training and compare two label-free alternatives that preserve the downstream \ac{FSL} architectures, thereby isolating the effects of the data regime.

Another important distinction is whether \ac{FSL} methods use frozen representations or adapt at test time (e.g., through \ac{IDA}). Some approaches learn generic features so that simple classifiers, such as nearest-neighbor or linear models, perform well on frozen embeddings~\cite{Tian:ECCV:2020, Dhillon:ICLR:2020}, whereas gradient-based methods such as \ac{MAML}~\cite{Chelsea:ICML:2017} incorporate support-set adaptation into training. Other studies show that even simple fine-tuning on $k$-shot support sets can yield substantial gains~\cite{Hu:IEEE:2022, Lin:ESwA:2023}. We isolate test-time adaptation from pre-training by evaluating each regime with and without episodic fine-tuning and analyzing how gains vary across \ac{OOD} sources. This separates improvements due to algorithmic changes from those driven by more favorable data regimes and benchmark choices.

\section{Method} 
\label{sec:method}

This section formalizes the few-shot setting considered in this work and the pre-training regimes under analysis. We first define the target and source domains, the episodic evaluation protocol, and the meta-risk used to compare training regimes. We then discuss how domain shift explains the optimistic bias introduced by standard \ac{ID} pre-training. Finally, we describe the proposed label-free pre-training strategies and the source-domain selection procedure.

\subsection{Problem formulation}
\label{sec:method:problem}

Let $\mathcal{X}$ and $\mathcal{Y}$ denote the input and label spaces, respectively. A domain is defined by a joint distribution $P(X,Y)$ over $\mathcal{X}\times\mathcal{Y}$. We distinguish the target domain $P_{\mathrm{tgt}}(X,Y)$, from which few-shot episodes are sampled, from the source domain $P_{\mathrm{src}}(X,Y)$ used for pre-training.

In few-shot classification, each episode $T=(S_T,Q_T)$ is defined by a support set $S_T$ and a query set $Q_T$, both constructed from a subset of target classes 
$\mathcal{C}_T \subset \mathcal{Y}_{\mathrm{tgt}}$, with $|\mathcal{C}_T|=n$ (i.e., $n$-way).
The support set contains $k$ labeled samples per class (i.e., $k$-shot),
\begin{equation}
S_T = \{(x_i,y_i)\}_{i=1}^{nk},
\quad
(x_i,y_i) \sim P_{\mathrm{tgt}}(X,Y \mid Y \in \mathcal{C}_T),
\label{eq:support_set}
\end{equation}
and, \textit{in principle}, should constitute the only labeled target-domain information available to the model within the episode. 
The query set $Q_T$ contains additional disjoint samples from the same classes in $C_T$ and is used to evaluate the episode performance.

We write a few-shot predictor for episode $T$ as
\begin{equation}
h_{\theta,T}(x) = g_T\big(f_\theta(x); S_T\big),
\label{eq:fsl_predictor}
\end{equation}

where $f_\theta$ is the learned representation and $\theta$ contains all trainable parameters shared across episodes. The function $g_T$ denotes the episode-conditioned classifier induced by the support set $S_T$, whose samples are also embedded through $f_\theta$ when performing comparisons. This general formulation is intended to cover metric-based methods such as Matching Networks, Prototypical Networks, and Relation Networks, which differ in how $g_T$ compares query and support embeddings. In methods with trainable comparison modules, such as Relation Networks, those shared parameters are treated as part of $\theta$.

\subsection{Pre-training regimes}
\label{sec:method:pt_regimes}

We compare three pre-training regimes, which differ in how the shared parameters $\theta$ are initialized before target-domain episodes are evaluated:

\begin{itemize}
    \item \textbf{No-PT:} the parameters $\theta$ are not initialized through source-domain pre-training. For each episode, $\theta$ is trained only on the scarce labeled examples in $S_T$.    

    \item \textbf{ID-PT:} the model is pre-trained on labeled classes from the same visual domain as the target data, but with disjoint label sets,
    $\mathcal{Y}_{\mathrm{src}} \cap \mathcal{Y}_{\mathrm{tgt}} = \varnothing$.
    This corresponds to the standard class-disjoint benchmark protocol used in many \ac{FSL} datasets.

    \item \textbf{OOD-PT:} the model is pre-trained on a source domain that differs from the target domain both in its label set and in its input distribution, i.e.,
    $\mathcal{Y}_{\mathrm{src}} \cap \mathcal{Y}_{\mathrm{tgt}} = \varnothing$
    and $P_{\mathrm{src}}(X) \neq P_{\mathrm{tgt}}(X)$.
\end{itemize}

Although ID-PT satisfies class disjointness, source and target samples are drawn from the same dataset or visual domain. Therefore, their input distributions are often highly similar, i.e., $P_{\mathrm{src}}(X)\approx P_{\mathrm{tgt}}(X)$, which makes the setting closer to an i.i.d. train/test split than to a strict low-data target-domain scenario. In contrast, OOD-PT introduces a domain shift between pre-training and evaluation, resulting in a more realistic---and more challenging---\ac{FSL} setting.

For the ID-PT and OOD-PT regimes, we also evaluate an \acf{IDA} step after pre-training. This step fine-tunes the model episodically using only the target support set $S_T$, which constitutes the only labeled target-domain data available at evaluation time. At each \ac{IDA} episode, class-balanced support and query subsets are sampled with replacement from the examples available in $S_T$ for each target class.

\subsection{Meta-risk and optimistic bias}
\label{sec:method:bias}

This section formalizes how different pre-training regimes affect the target meta-risk\footnote{The target meta-risk measures the expected query-set loss on a previously unseen target task, represented by an episode, after conditioning the predictor on that episode's support set under the fixed $n$-way $k$-shot protocol.} under the same few-shot evaluation protocol, and how this can induce an optimistic bias when ID pre-training is performed. All quantities used in the formulation below are defined for a fixed $n$-way $k$-shot setting, whose dependence is omitted for clarity. For a loss function $\ell$, the target meta-risk of parameters $\theta$ is defined as
\begin{equation}
\mathcal{R}_{\mathrm{tgt}}(\theta)
=
\mathbb{E}_{T\sim p_{\mathrm{tgt}}}
\mathbb{E}_{(x,y)\sim Q_T}
\left[
\ell\big(h_{\theta,T}(x),y\big)
\right],
\label{eq:target_metarisk}
\end{equation}
where $p_{\mathrm{tgt}}$ denotes the episodic distribution induced by $P_{\mathrm{tgt}}(X,Y)$ under the fixed $n$-way $k$-shot protocol, $T$ denotes a target episode with support set $S_T$ and query set $Q_T$, $(x,y)\sim Q_T$ denotes uniform averaging over query samples, and $h_{\theta,T}$ is the support-conditioned predictor induced by the model parameters $\theta$. Analogously, $\mathcal{R}_{\mathrm{src}}(\theta)$ denotes the expected meta-risk over source-domain episodes.

The effect of the pre-training domain can be interpreted through the lens of domain adaptation theory~\cite{BenDavid:MLJ:2010}. Although classical domain-adaptation bounds are not directly formulated for class-disjoint episodic few-shot learning, they provide a useful decomposition of the factors that affect transfer from a source domain to a target domain. In particular, we consider the following adaptation-inspired relation:
\begin{equation}
\begin{aligned}
\mathcal{R}_{\mathrm{tgt}}(\theta)
\;\lesssim\;&
\mathcal{R}_{\mathrm{src}}(\theta)
+
\tfrac{1}{2}
\mathrm{disc}_{\mathcal{H}}\bigl(
f_\theta\#P_{\mathrm{src}}(X),
f_\theta\#P_{\mathrm{tgt}}(X)
\bigr) \\[0.2em]
&{}
+
\lambda_{\mathrm{src},\mathrm{tgt}}^\theta .
\end{aligned}
\label{eq:domain_bound}
\end{equation}
where $f_\theta\#P(X)$ denotes the embedding distribution induced by the encoder $f_\theta$, $\mathrm{disc}_{\mathcal{H}}$ measures the source--target discrepancy in representation space under a hypothesis class $\mathcal{H}$, and $\lambda_{\mathrm{src},\mathrm{tgt}}^\theta$ captures the residual joint error of the best support-conditioned predictor for both domains. 
In this relation, these terms play different roles in our setting:

\begin{itemize}
    \item The source risk $\mathcal{R}_{\mathrm{src}}(\theta)$ reflects how well the model solves source-domain episodes. By itself, it does not encode whether the source is visually aligned with the target.
    \item The discrepancy term $\mathrm{disc}_{\mathcal{H}}$ captures the relation between the pre-training and target domains. In ID-PT, this term is expected to be small. In OOD-PT, the source and target input distributions may differ substantially, making this term larger and source-dependent. 
    \item The residual term $\lambda_{\mathrm{src},\mathrm{tgt}}^\theta$ captures the remaining source--target incompatibility under the representation and support-conditioned predictor. The \ac{IDA} step can reduce this mismatch by adapting the model to $S_T$.
\end{itemize}

This decomposition highlights the difference between class disjointness and domain disjointness. Standard ID-PT satisfies $\mathcal{Y}_{\mathrm{src}}\cap\mathcal{Y}_{\mathrm{tgt}}=\varnothing$, but it still exposes the representation to samples that are visually close to those later appearing in the support and query sets. Consequently, ID-PT can reduce the target meta-risk by aligning the representation with the target visual distribution before few-shot evaluation. This induces an optimistic evaluation regime.

Let $\theta_m$ denote the model parameters obtained after pre-training on source domain $m$, and let $\theta_{\varnothing}$ denote the No-PT model. We define the risk-based transfer gain of source $m$ on target domain $\mathrm{tgt}$ as
\begin{equation}
\Delta_{\mathcal{R}}(m,\mathrm{tgt})
=
\mathcal{R}_{\mathrm{tgt}}(\theta_{\varnothing})
-
\mathcal{R}_{\mathrm{tgt}}(\theta_m).
\label{eq:delta_risk}
\end{equation}
Positive values of $\Delta_{\mathcal{R}}(m,\mathrm{tgt})$ indicate that pre-training on source $m$ reduces the target meta-risk with respect to No-PT, whereas negative values indicate detrimental transfer.

For ID-PT, let $m_{\mathrm{ID}}$ denote an ID source. For OOD-PT, let $m_{\mathrm{OOD}}$ denote an OOD source. We define the optimistic bias induced by ID-PT as the excess transfer gain obtained from the ID source:
\begin{equation}
B_{\mathcal{R}}(m_{\mathrm{ID}},m_{\mathrm{OOD}},\mathrm{tgt})
=
\Delta_{\mathcal{R}}(m_{\mathrm{ID}},\mathrm{tgt})
-
\Delta_{\mathcal{R}}(m_{\mathrm{OOD}},\mathrm{tgt}).
\label{eq:optimistic_bias}
\end{equation}
A positive $B_{\mathcal{R}}$ indicates that ID-PT provides an additional advantage over OOD-PT. According to Eq.~\eqref{eq:domain_bound}, this advantage is consistent with the smaller source--target discrepancy expected in representation space when the representation is pre-trained on a distribution that is already visually close to the target evaluation episodes.

For ID-PT and OOD-PT, we also quantify the gain provided by \ac{IDA}. Let $\theta_m^{\mathrm{IDA}}$ denote the episode-adapted parameters obtained by applying IDA to $\theta_m$ using $S_T$ before evaluating the corresponding query set. The risk-based contribution of IDA is
\begin{equation}
\Delta_{\mathcal{R}}^{\mathrm{IDA}}(m,\mathrm{tgt})
=
\mathcal{R}_{\mathrm{tgt}}(\theta_m)
-
\mathcal{R}_{\mathrm{tgt}}(\theta_m^{\mathrm{IDA}}).
\label{eq:ida_gain}
\end{equation}
If $\Delta_{\mathcal{R}}^{\mathrm{IDA}}(m_{\mathrm{OOD}},\mathrm{tgt})$ is larger than $\Delta_{\mathcal{R}}^{\mathrm{IDA}}(m_{\mathrm{ID}},\mathrm{tgt})$, this indicates that the support set contributes more to target adaptation when the representation has not already been aligned with the target domain during pre-training. Conversely, a small or negative IDA gain under ID-PT suggests that most of the improvement has already been obtained through ID pre-training, leaving little additional benefit for adaptation based only on the available $n\times k$ labeled target samples.

\subsection{Label-free \ac{OOD} pre-training}
\label{sec:method:labelfree}

As discussed before, one way to mitigate the optimistic bias introduced by \ac{ID} pre-training is to rely on \ac{OOD} data instead. However, even when a suitable \ac{OOD} source is available for a given target dataset, standard pre-training assumes that this external source is fully labeled. This requirement can be restrictive in practice and may significantly reduce the set of candidate source datasets. To remove this dependency on source annotations, we propose two label-free pre-training strategies that construct episodic supervision directly from unlabeled samples $x \sim P_{\mathrm{src}}(X)$.

\subsubsection{Clustering-based pseudo-labels}
\label{sec:method:uckm}

Our first label-free \ac{OOD} pre-training strategy, which we denote as \ac{UCKM}, constructs pseudo-labels by clustering the unlabeled source dataset and treating each cluster as a class. Concretely, we partition the source dataset into $K_c$ clusters using visual features extracted from its images. This partition defines a pseudo-label space $\tilde{\mathcal{Y}}=\{1, \dots, K_c\}$, where each sample is assigned the index of its cluster.

\begin{figure}[ht]
    \centering
    \includegraphics[width=0.6\columnwidth]{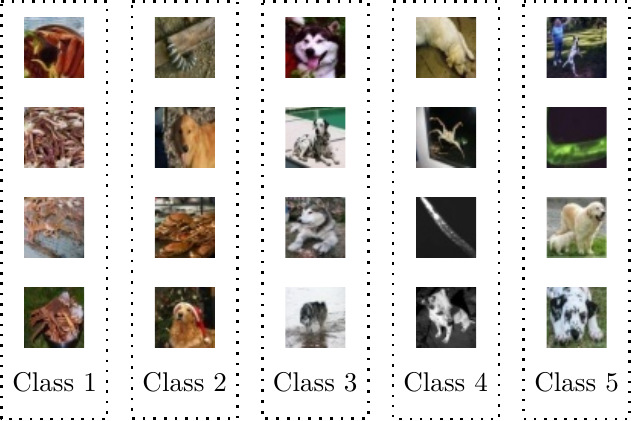}
    \caption{Examples of the pseudo-labeling process using \ac{UCKM} on the miniImageNet dataset.}
    \label{fig:exampleUCKM}
\end{figure}

Since clustering is performed directly from image features, the induced pseudo-classes do not necessarily correspond to human-interpretable categories. Instead, samples may be grouped according to regularities in the source domain such as texture, color, or other latent visual attributes (see Figure~\ref{fig:exampleUCKM} for an example). Nevertheless, these pseudo-classes provide a useful supervisory signal for episodic pre-training by encouraging the encoder to separate distinct visual groups.

For episodic pre-training to be feasible, each cluster must contain enough samples to form both the support and query sets. In an $n$-way $k$-shot setting with $q$ query samples per class, this requires at least $k+q$ samples per cluster. To enforce this constraint, we use the constrained K-means algorithm of Bradley et al.~\cite{bradley2000constrained}, which allows setting lower bounds on cluster size and thus prevents the creation of too small clusters.

After clustering, each source sample is assigned its cluster index as a pseudo-label, and episodic pre-training proceeds in the same way as in the supervised setting.

The number of clusters $K_c$ controls the granularity of the partition: smaller values produce coarser pseudo-classes, whereas larger values yield finer groupings that may better capture the diversity of the source data. We therefore treat $K_c$ as a hyperparameter and study its effect empirically.

\subsubsection{Instance-based pseudo-labeling via augmentations}
\label{sec:method:uiaug}

Our second label-free strategy, referred to as \ac{UIAug}, is inspired by the instance discrimination paradigm introduced in self-supervised representation learning \cite{Wu2018instdisc,Chen2020simclr,Jiang:Interspeech:2021}. In these approaches, different augmented views of the same image are treated as positive pairs, while views from different images are treated as negatives, typically under contrastive objectives. Instead of relying on pairwise comparisons between augmented samples, we adapt this principle to the episodic meta-learning setting by turning instance discrimination into a synthetic $n$-class classification task.

Specifically, each unlabeled image $x_i$ from the source dataset is treated as a distinct pseudo-class with identifier $\tilde{y}_i$.
For each episode, we sample $n$ images from the unlabeled source dataset, which define the $n$ pseudo-classes of the episode. 
For each selected image $x_i$, we generate $k$ stochastic augmentations
$\{\tilde{x}_i^{(1)}, \ldots, \tilde{x}_i^{(k)}\}$ 
using standard image transformations (e.g., rotations, translations, scaling, blurring, or color perturbations). These augmented samples populate the support set 
$\tilde{S}_T = \{(\tilde{x}_i^{(j)}, \tilde{y}_i) \mid i=1,\dots,n,\; j=1,\dots,k\}$,
while the original, non-augmented images are used to form the query set 
$\tilde{Q}_T = \{(x_i, \tilde{y}_i) \mid i=1,\dots,n\}$.

\begin{figure}[ht]
    \centering
    \includegraphics[width=0.6\columnwidth]{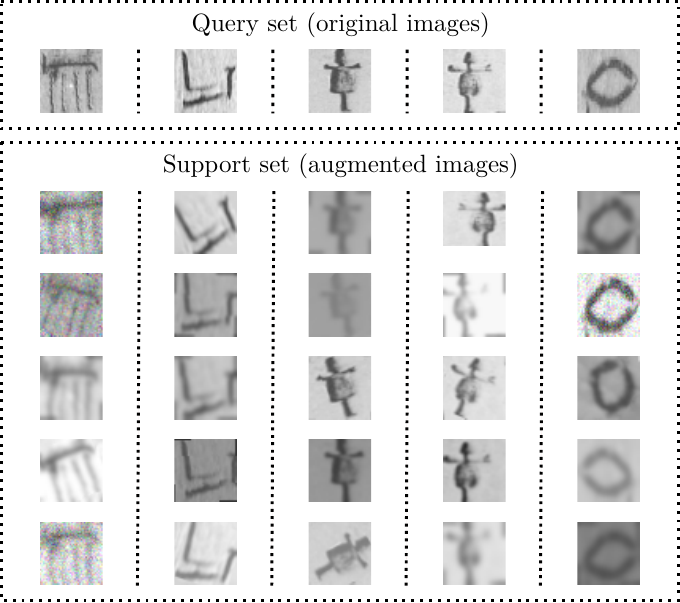}
    \caption{\ac{UIAug} construction for a 5-way 5-shot episode. Each column corresponds to a pseudo-class defined by a single source instance, whose augmentations form the support set.}
    \label{fig:exampleUIAug}
\end{figure}

This construction preserves the standard $n$-way $k$-shot episodic structure while deriving supervision directly from unlabeled data. Figure~\ref{fig:exampleUIAug} illustrates the episode construction for a 5-way 5-shot example.

Unlike the clustering-based strategy, which groups different images into pseudo-classes according to their visual regularities, \ac{UIAug} generates multiple transformed views of the same image and treats each original instance as its own pseudo-class. In addition, \ac{UIAug} does not require an explicit clustering stage or the specification of the number of clusters $K_c$, making it a simple and scalable approach.

\subsection{Source-domain selection prior to pre-training}
\label{sec:method:source_selection}

When multiple candidate \ac{OOD} datasets are available, selecting the most suitable pre-training domain can substantially affect downstream performance (as shown in Fig.~\ref{fig:intro_example}). In practice, however, identifying the best option is non-trivial, as it would require exhaustively pre-training and evaluating a separate model for each candidate source, which is computationally prohibitive. To address this limitation, we propose a \emph{source-selection} procedure that ranks candidates for a given target using \emph{domain-similarity} signals that can be computed \emph{before} pre-training.

Let $\mathcal{S}=\{P_{\mathrm{src}}^{(1)},\ldots,P_{\mathrm{src}}^{(M)}\}$ be a pool of $M$ candidate source domains and $P_{\mathrm{tgt}}$ a fixed target domain. Ideally, source selection would choose the domain $m$ that maximizes the transfer gain $\Delta_{\mathcal{R}}(m,\mathrm{tgt})$, as defined in Eq.~\eqref{eq:delta_risk}. This gain, however, is only revealed after learning source-specific parameters $\theta$ and evaluating them on the target domain, making it unavailable before pre-training. We therefore propose to approximate this oracle choice using a scoring function computed before pre-training:
\begin{equation}
\hat{m}
=
\argmax_{m\in\{1,\ldots,M\}}
\mathrm{Score}\big(P_{\mathrm{src}}^{(m)},P_{\mathrm{tgt}}\big).
\label{eq:argmax_score}
\end{equation}
To instantiate $\mathrm{Score}(\cdot,\cdot)$, we consider two strategies: individual similarity descriptors and a composite score based on factor analysis, as detailed below.

\subsubsection{Individual-descriptor scoring}

We consider a set of $J$ individual descriptors that quantify source--target similarity from different perspectives, including dataset-level metadata (e.g., number of classes, number of samples, or samples per class), low-level image statistics (e.g., brightness, contrast, sharpness, or colorfulness), representation-space overlap, prediction-based distance and compactness patterns, and reconstruction-based similarity. A detailed description of these descriptors is provided in Appendix~\ref{app:similarity-metrics}.

For each source--target pair $(P_{\mathrm{src}}^{(m)}, P_{\mathrm{tgt}})$, these $J$ descriptors are arranged into a similarity vector
$\mathbf{s}_{m,\mathrm{tgt}} =
[s^{(1)}_{m,\mathrm{tgt}}, \ldots, s^{(J)}_{m,\mathrm{tgt}}].$ 
All descriptors are represented numerically and standardized. 
In the individual-descriptor setting, each descriptor $j$ defines its own scoring function: 
$
\mathrm{Score}_j\big(P_{\mathrm{src}}^{(m)}, P_{\mathrm{tgt}}\big) = s^{(j)}_{m,\mathrm{tgt}}.
$

To assess how informative each descriptor is for source selection, we perform an offline calibration stage in which the transfer gain $\Delta_{\mathcal R}(m,\mathrm{tgt})$ is estimated for a collection of source--target pairs by running the full pre-training and target-evaluation pipeline. Using these observed gains, we quantify the association between each descriptor and downstream transfer through the Pearson correlation, computed over all considered source--target pairs $(m,\mathrm{tgt})$:
\begin{equation}
r^{(j)} = \mathrm{Corr}\!\left(s^{(j)}_{m,\mathrm{tgt}}, \Delta_{\mathcal R}(m,\mathrm{tgt})\right).
\label{eq:pearson_descriptor_gain}
\end{equation}

The magnitude of $r^{(j)}$ reflects how strongly descriptor $j$ is associated with downstream few-shot gains. Once this calibration is fixed, source selection for a new target only requires computing the corresponding descriptor values, without exhaustively pre-training on all candidate sources.

\subsubsection{Factor-analysis composite scoring}
\label{sec:method:factor_analysis}

Individual descriptors may be noisy, redundant, or only partially informative when considered in isolation. We therefore construct a composite score by applying factor analysis to the standardized similarity vectors defined above. 
Concretely, the vectors $\mathbf{s}_{m,\mathrm{tgt}}$ computed for all calibration source--target pairs are stacked into a descriptor matrix whose rows correspond to source--target pairs and whose columns correspond to the $J$ standardized descriptors. Before factor extraction, descriptors with low individual Kaiser--Meyer--Olkin (KMO) adequacy are removed~\cite{Shrestha:AJAMS:2021}. The number of factors is selected using the Kaiser rule, retaining factors with eigenvalues greater than or equal to one.

Factor analysis is then applied to the retained descriptors. We evaluate different rotation configurations, including standard rotations and, during calibration, a target-aligned Procrustes rotation. Let $\mathcal{F}$ denote the set of resulting factor scores, and let $z_f$ be the score associated with factor $f$. We select the factor most strongly associated with the observed transfer gains $\Delta_{\mathcal R}$:

\begin{equation}
f^\star
=
\arg\max_{f\in\mathcal{F}}
\left|
\mathrm{Corr}\big(z_f,\Delta_{\mathcal R}\big)
\right|.
\label{eq:best_factor}
\end{equation}

The final factor-analysis score $z_{f^\star}$ is obtained from the selected factor $f^\star$ after sign alignment according to its correlation with $\Delta_{\mathcal R}$, so that larger values indicate higher expected transfer gain. This score is then used as $\mathrm{Score}(\cdot,\cdot)$ in Eq.~\eqref{eq:argmax_score} to rank candidate sources.

\section{Experimental setup} 
\label{sec:setup}

We evaluate the effect of pre-training under controlled few-shot conditions by systematically varying the dataset, source--target domain pair, model architecture, episodic configuration, and adaptation regime. This section details the experimental setup used for the empirical evaluation. The source code and configuration files required to reproduce all experiments are publicly available at \url{https://github.com/Alejandro-Galan/fsl_critique}.

\subsection{Datasets}
\label{sec:exp:dataset}

We use eight image classification datasets spanning heterogeneous visual domains, formats, and class distributions, summarized in Table~\ref{tab:datasets}. They include music notation (\CAPITAN{}~\cite{AlfaroContreras:PRL:2023}); handwritten characters and symbols (\Greek{}~\cite{Gatos2015GreekDataset}, \Egyptian{}~\cite{Franken2013EgyptianDataset}, \TKH{}~\cite{Yang2018TKHdataset}, and \Omniglot{}~\cite{Lake:AAAS:2015}); medical imaging (\OrganAMNIST{} from MedMNIST~\cite{Yang:ScientificData:2023}); and natural images (\MiniImageNet{}~\cite{Vinyals:NeurIPS:2016} and \CIFAR{}~\cite{Bertinetto:arxiv:2019}). This selection provides a broad set of source--target domain relationships, ranging from closely related symbolic domains to markedly different ones, enabling analysis of how pre-training is affected by differences in semantic content, visual appearance, class count, and imbalance.

\begin{table}[ht]
    \caption{Datasets used in the experimental evaluation. The table reports, for each dataset, its acronym, semantic domain, color mode, original image resolution (px.), number of classes, total number of samples, and samples-per-class statistics (minimum, maximum, and mean number of samples per class).}
    \label{tab:datasets}
    \centering
    \resizebox{\textwidth}{!}{%
            \begin{tabular}{llcccrrr}
                \toprule
                \textbf{Dataset} &
                \textbf{Acr.} &
                \textbf{Semantic domain} &
                \begin{tabular}[c]{@{}c@{}}\textbf{Color}\\\textbf{mode}\end{tabular} &
                \begin{tabular}[c]{@{}c@{}}\textbf{Original}\\\textbf{resolution}\end{tabular} &
                \textbf{Classes} &
                \begin{tabular}[c]{@{}r@{}}\textbf{Total}\\\textbf{samples}\end{tabular} &
                \begin{tabular}[c]{@{}c@{}}\textbf{Samples per class}\\\textbf{(min, max, mean)}\end{tabular} \\
                
                \midrule        
                
                \textbf{\CAPITAN{}}~\cite{AlfaroContreras:PRL:2023} &
                CAP. &
                Music notation &
                RGB &
                $64\times64$ &
                28 &
                $17\,112$ &
                $(57, 2\,737, 596)$ \\
        
                \textbf{\Egyptian{}}~\cite{Franken2013EgyptianDataset} &
                Egypt. &
                Historical symbols &
                Gray &
                $64\times64$ &
                18 &
                $4\,210$ &
                $(49, 447, 163)$ \\
        
                \textbf{\Greek{}}~\cite{Gatos2015GreekDataset} &
                Greek &
                Handwritten characters &
                RGB &
                $64\times64$ &
                112 &
                $168\,719$ &
                $(50, 12\,725, 1\,494)$ \\     
        
                \textbf{\Omniglot{}}~\cite{Lake:AAAS:2015} &
                Omni. &
                Handwritten characters &
                Gray &
                $105\times105$ &
                1\,623 &
                $32\,460$ &
                $(20, 20, 20)$ \\
        
                \textbf{\TKH{}}~\cite{Yang2018TKHdataset} &
                TKH &
                Handwritten characters &
                RGB &
                $64\times64$ &
                347 &
                $323\,498$ &
                $(49, 16\,812, 908)$ \\
        
                \textbf{\OrganAMNIST{}}~\cite{Yang:ScientificData:2023} &
                oMNIST &
                Medical imaging &
                Gray &
                $64\times64$ &
                11 &
                $64\,932$ &
                $(2\,407, 10\,482, 5\,348)$ \\       
        
                \textbf{\CIFAR{}}~\cite{Bertinetto:arxiv:2019} &
                CIF. &
                Natural images &
                RGB &
                $32\times32$ &
                100 &
                $60\,000$ &
                $(600, 600, 600)$ \\        
        
                \textbf{\MiniImageNet{}}~\cite{Vinyals:NeurIPS:2016} &
                mINet &
                Natural images &
                RGB &
                $84\times84$ &
                100 &
                $60\,000$ &
                $(600, 600, 600)$ \\
                
                \bottomrule
            \end{tabular}
    }
\end{table}

For \MiniImageNet{}, \CIFAR{}, and \Omniglot{}, we use the public class-disjoint splits introduced in the original publications. For the remaining datasets, whose original partitions were not designed for class-disjoint meta-learning evaluation, we generate a new random class-disjoint split in each bootstrap run, assigning approximately $80\%$ of the classes to meta-training and $20\%$ to meta-testing. Thus, all target-domain results are evaluated on classes unseen during pre-training.

Since our experimental design compares all datasets within a common episodic framework and treats each domain as both a potential pre-training source and a target, all images are rescaled to a common spatial resolution of $40\times40$ pixels and normalized to $[0,1]$ by dividing pixel intensities by 255. This resolution limits interpolation artifacts in low-resolution datasets such as \CIFAR{} while preserving the visual structure of higher-resolution images after downsampling. The standardized input enables the same architecture to be applied across all source--target pairs and prevents transfer performance differences from being attributed to resolution or intensity scaling. Accordingly, our results are intended for the proposed cross-domain comparison and may not be directly comparable to benchmarks using the original resolutions or preprocessing pipelines.

\subsection{Implementation details}
\label{sec:exp:implementation}

We evaluate three representative metric-based meta-learning architectures: Matching Networks~\cite{Vinyals:NeurIPS:2016}, Prototypical Networks~\cite{Snell:NeurIPS:2017}, and Relation Networks~\cite{Sung:CVPR:2018}. These models cover the main families of episodic similarity-based classification methods and provide a common basis for comparing the proposed pre-training regimes.

We follow standard few-shot evaluation settings and consider $5$- and $20$-way classification with $1$, $5$, and $10$ shots, evaluating all feasible combinations across architectures, datasets, and pre-training regimes.\footnote{A dataset is eligible as a pre-training source or target in $20$-way experiments only if its meta-training or meta-test split, respectively, contains at least 20 classes.} Each experiment comprises 4\,000 test episodes, while pre-training, when applicable, also runs for 4\,000 episodes. Results are averaged over five bootstrap runs with different random seeds and class-disjoint partitions. All compared methods are evaluated on identical experiment sets, and we report both overall averages and results disaggregated by way, shot, architecture, and target domain.

Following~\cite{Snell:NeurIPS:2017}, we use an initial learning rate of $10^{-3}$, a weight decay of $10^{-6}$, and a learning-rate step size of 2\,000 episodes. For \ac{IDA}, adaptation uses only the labeled support set of each target episode and consists of three 1\,000-episode stages with a learning rate of $10^{-4}$.

For \ac{UIAug}, we use random resized crops with scales in $[0.85, 1.0]$, rotations within $\pm5^\circ$, color jittering with probability 0.8, gray-scale conversion with probability 0.2, Gaussian blur, and additive Gaussian noise. Flips are excluded because orientation may be semantically relevant in medical images, alphabets, and music notation.

\subsection{Evaluation metrics}
\label{sec:exp:metrics}

For experimental evaluation, we report mean query accuracy, the standard metric in few-shot classification, together with its standard deviation across bootstrap runs. Although the methodological formulation is expressed in terms of meta-risk, accuracy provides an empirical counterpart under the 0--1 classification setting. Since our goal is to quantify the effect of each pre-training regime, we report accuracy gains with respect to the No-PT baseline. Let $\mathrm{Acc}(m,\mathrm{tgt})$ be the mean query accuracy on target domain $\mathrm{tgt}$ after pre-training on source domain $m$, and let $\mathrm{Acc}(\varnothing,\mathrm{tgt})$ denote the corresponding No-PT accuracy. The accuracy gain of pre-training on source $m$ over No-PT is
\begin{equation}
\Delta_{\mathrm{Acc}}(m,\mathrm{tgt})
=
\mathrm{Acc}(m,\mathrm{tgt})
-
\mathrm{Acc}(\varnothing,\mathrm{tgt}), 
\label{eq:delta_acc}
\end{equation}
where positive values indicate beneficial transfer, whereas negative values indicate detrimental transfer.

The effect of \ac{IDA} is measured as
\begin{equation}
\Delta_{\mathrm{Acc}}^{\mathrm{IDA}}(m,\mathrm{tgt})
=
\mathrm{Acc}^{\mathrm{IDA}}(m,\mathrm{tgt})
-
\mathrm{Acc}(m,\mathrm{tgt}),
\label{eq:ida_acc_gain}
\end{equation}
where $\mathrm{Acc}^{\mathrm{IDA}}(m,\mathrm{tgt})$ is the mean query accuracy after episodic adaptation on the target support set. This difference isolates the gain obtained from the only labeled target samples available at evaluation time.

\section{Results}
\label{sec:results}

This section presents the experimental results from two complementary perspectives. First, we evaluate how different pre-training regimes affect few-shot performance, comparing ID-PT, supervised OOD-PT, and the proposed label-free OOD methods. Second, we examine source-domain selection in OOD-PT by analyzing how source--target relationships influence transfer and whether domain-level metrics can identify suitable pre-training sources.

\subsection{Effect of Pre-Training on Few-Shot Performance}

We begin by assessing the role of pre-training through $\Delta_{\mathrm{Acc}}$ (Eq.~\ref{eq:delta_acc}), using No-PT as a common reference. This highlights the gap between ID-PT and OOD-PT, the extent to which IDA mitigates source--target mismatch, and the effectiveness of the proposed label-free OOD-PT variants.

\subsubsection{ID-PT vs. OOD-PT}

\begin{figure}[ht]
    \centering
    \includegraphics[width=0.7\columnwidth]{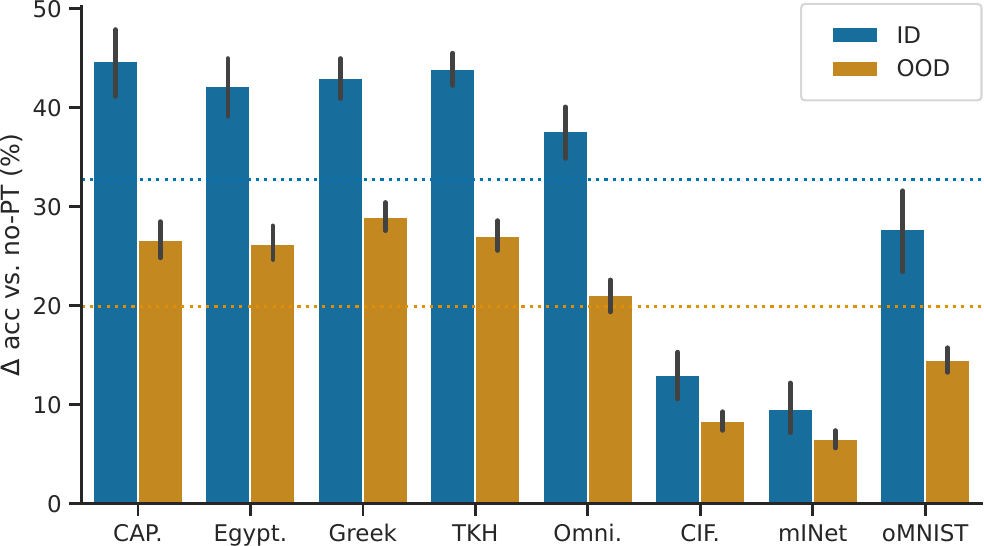}
    \caption{Accuracy gain over No-PT for ID-PT and supervised OOD-PT. Bars report $\Delta_{\mathrm{Acc}}$ for each target dataset, averaged over architectures, bootstrap runs, $n$-way configurations, and $k$-shot settings. Error bars indicate the variability across the averaged experimental configurations, and dotted lines denote the overall mean gain for each pre-training strategy.}
    \label{fig:results:ID}
\end{figure}

Figure~\ref{fig:results:ID} reports $\Delta_{\mathrm{Acc}}$ for ID-PT and supervised OOD-PT before IDA. In ID-PT, source and target domains coincide while preserving class-disjoint pre-training and evaluation splits. In supervised OOD-PT, results are averaged over all source domains different from the target. Error bars reflect variability across experimental configurations, including, in the OOD-PT case, the variability induced by the choice of source domain.

Both pre-training regimes outperform No-PT, but by different margins. ID-PT yields the largest gains, averaging 33.41 percentage points (pp.) across target datasets, compared with 23.75 pp. for supervised OOD-PT. This 9.66 pp. gap shows that class-disjoint pre-training within the target domain provides a substantial advantage beyond the general benefit of pre-training.

The gap is most pronounced in domain-specific datasets. For \CAPITAN{}, \Egyptian{}, \Greek{}, \TKH{}, and \Omniglot{}, ID-PT improves accuracy by approximately 37--46 pp., consistently exceeding supervised OOD-PT. The advantage is smaller for the natural-image datasets, where ID-PT yields gains of 13.43 pp. on \CIFAR{} and 9.97 pp. on \MiniImageNet{}, with gaps over supervised OOD-PT of 2.98 and 2.02 pp., respectively. These results support our central claim that ID-PT provides an optimistic evaluation setting for FSL, whereas OOD-PT offers a more conservative and realistic performance estimate.

\begin{figure}[ht]
    \centering
    \begin{subfigure}[c]{0.48\textwidth}
        \centering
        \includegraphics[width=1.\linewidth]{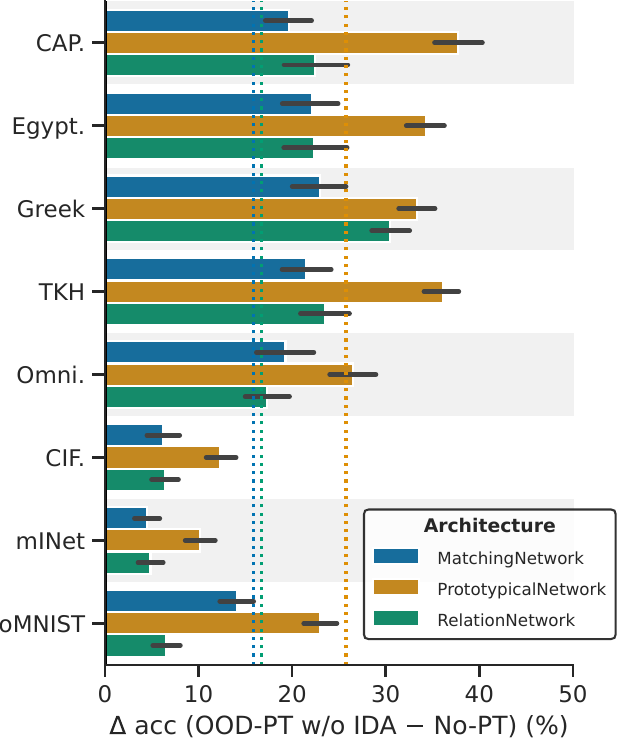}
        \caption{Grouped by architecture.}
         \label{fig:results:OOD:Supervised:generalResults:model}
    \end{subfigure}
    \begin{subfigure}[c]{0.46\textwidth}
        \centering
        \includegraphics[width=1.\linewidth]{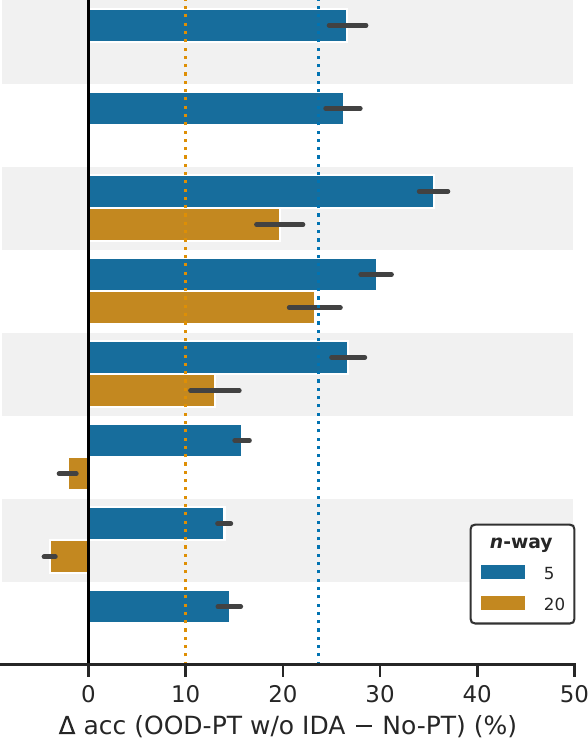}
        \caption{Grouped by \textit{N}-way.}
        \label{fig:results:OOD:Supervised:generalResults:spc}
    \end{subfigure}
    \caption{Mean accuracy gain of supervised OOD-PT over No-PT for each target dataset, averaged across all OOD source domains. Results are grouped by architecture in \textbf{(a)} and by $n$-way configuration in \textbf{(b)}. Bars indicate mean $\Delta_{\mathrm{Acc}}$, error bars show variability across source domains, and dotted lines mark the overall mean across targets.}
    \label{fig:results:OOD}
\end{figure}

Figure~\ref{fig:results:OOD} further analyzes supervised OOD-PT by architecture and $n$-way configuration. Prototypical Networks consistently obtain the largest gains, exceeding 30 pp. on several domain-specific datasets, whereas Matching Networks and Relation Networks show lower and more similar improvements. One possible explanation is that prototype-based classification transfers more robustly across domains because it relies on class-level representations rather than learning a more flexible similarity function, which may require more target-specific evidence and be more sensitive to noise or outliers.

The $n$-way analysis shows that the benefit of supervised OOD-PT decreases as the task becomes more demanding. In the 5-way setting, it improves performance for all target datasets, with an average gain of 23.64 pp. and larger gains on domain-specific datasets. In the 20-way setting, the average gain remains positive at 10 pp., but the effect becomes more target-dependent. OOD-PT still improves performance on symbolic datasets, while yielding negative gains on the natural-image datasets. This suggests that increasing the number of target classes makes OOD transfer less reliable when source and target domains are less aligned.

\subsubsection{Effect of IDA}

\begin{figure}[ht]  
    \centering
    \begin{subfigure}[c]{0.46\columnwidth}
        \includegraphics[width=\textwidth]{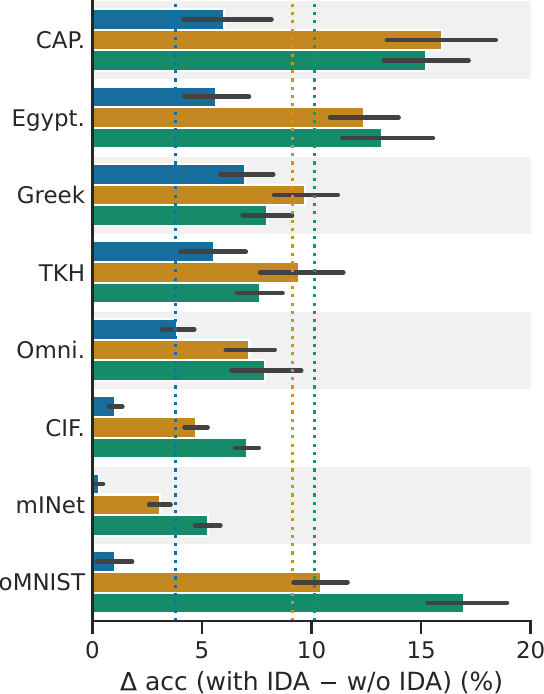}
        \caption{Supervised OOD-PT.} 
        \label{fig:results:IDA:ODD}
    \end{subfigure}
    \begin{subfigure}[c]{0.46\columnwidth}
        \includegraphics[width=\textwidth]{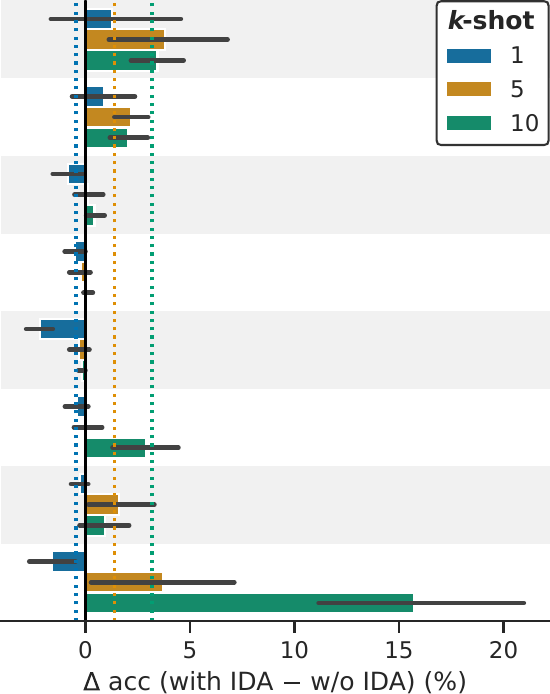}
        \caption{ID-PT.} 
        \label{fig:results:IDA:ID}
    \end{subfigure}
    \caption{Effect of \ac{IDA} under \textbf{(a)} supervised OOD-PT and \textbf{(b)} ID-PT, measured by $\Delta_{\mathrm{Acc}}^{\mathrm{IDA}}$ (Eq.~\ref{eq:ida_acc_gain}) for each target dataset and $k$-shot setting. Bars show mean gains, error bars indicate variability across experimental configurations, and dotted lines mark the overall mean for each $k$-shot setting.}
    \label{fig:results:IDA}
\end{figure}

We next examine the effect of \ac{IDA} after pre-training. Figure~\ref{fig:results:IDA} reports the adaptation gain, $\Delta_{\mathrm{Acc}}^{\mathrm{IDA}}$ (Eq.~\ref{eq:ida_acc_gain}), for supervised OOD-PT and ID-PT across $k$-shot settings.

The main observation is that the effect of \ac{IDA} is noticeably stronger under OOD-PT. It improves performance for every target dataset and $k$-shot setting, with an average gain of 7.69 pp. The improvement increases with the number of support samples, from 3.80 pp. in 1-shot to 9.12 pp. in 5-shot and 10.15 pp. in 10-shot. The largest gains again occur mainly in domain-specific datasets, with \OrganAMNIST{} reaching 16.97 pp. in 10-shot and \CAPITAN{} exceeding 15 pp. in both 5-shot and 10-shot. These results indicate that \ac{IDA} effectively uses the available target support set to reduce source--target mismatch.

Under ID-PT, the benefit of \ac{IDA} is substantially smaller, with an average gain of 1.37 pp. The 1-shot setting yields a negative average effect of $-0.45$ pp., with several datasets showing negligible or adverse gains. This suggests that, when pre-training is performed on the target domain, adaptation provides limited additional information and may even overfit to the small support set. Although the effect increases with $k$, reaching 3.19 pp. in 10-shot, it remains substantially below the gains observed under OOD-PT.

\subsubsection{Label-Free OOD-PT}

This section evaluates OOD pre-training when source-domain labels are not available using the two strategies introduced in Section~\ref{sec:method:labelfree}: \ac{UCKM} and \ac{UIAug}.

\paragraph*{\ac{UCKM}}

\begin{figure}[ht]
    \centering
    \includegraphics[width=0.8\columnwidth]{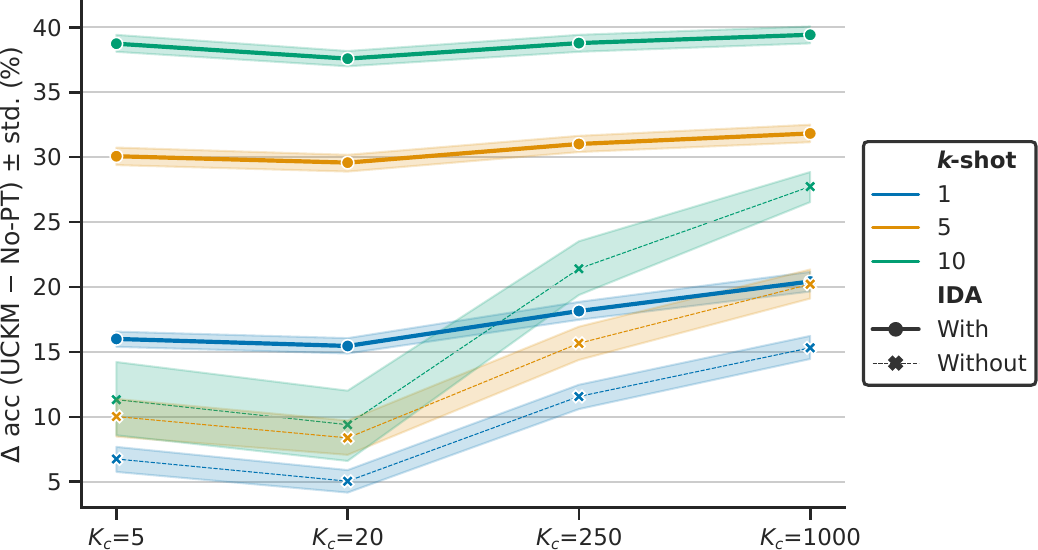}
    \caption{Effect of the number of clusters in \ac{UCKM}, reported as $\Delta_{\mathrm{acc}}$ for $K_c \in \{5,20,250,1000\}$ across $k$-shot settings with and without \ac{IDA}. Lines show the mean gain over the No-PT baseline across experimental configurations.}
    \label{fig:results:UCKM}
\end{figure}

Figure~\ref{fig:results:UCKM} analyzes the effect of the number of clusters in \ac{UCKM}, with $K_c \in \{5, 20, 250, 1000\}$, using $\Delta_{\mathrm{acc}}$ in settings with and without \ac{IDA}. Overall, increasing the number of clusters provides more informative supervision during pre-training, with $K_c=1000$ yielding the largest gains across all $k$-shot settings and both \ac{IDA} configurations.

The effect of $K_c$ is most pronounced without \ac{IDA}, where the gap between the worst and best configurations reaches 10.16, 11.73, and 18.34 pp. for 1-, 5-, and 10-shot, respectively. With \ac{IDA}, these gaps decrease to 4.80, 2.21, and 1.84 pp., indicating that adaptation reduces sensitivity to the number of clusters and partially compensates for less informative pseudo-label structures. Since $K_c=1000$ still performs best in all settings, we use it in the remaining \ac{UCKM} experiments.

\paragraph*{\ac{UIAug}}

\begin{figure}[ht]
    \centering
    \includegraphics[width=0.6\columnwidth]{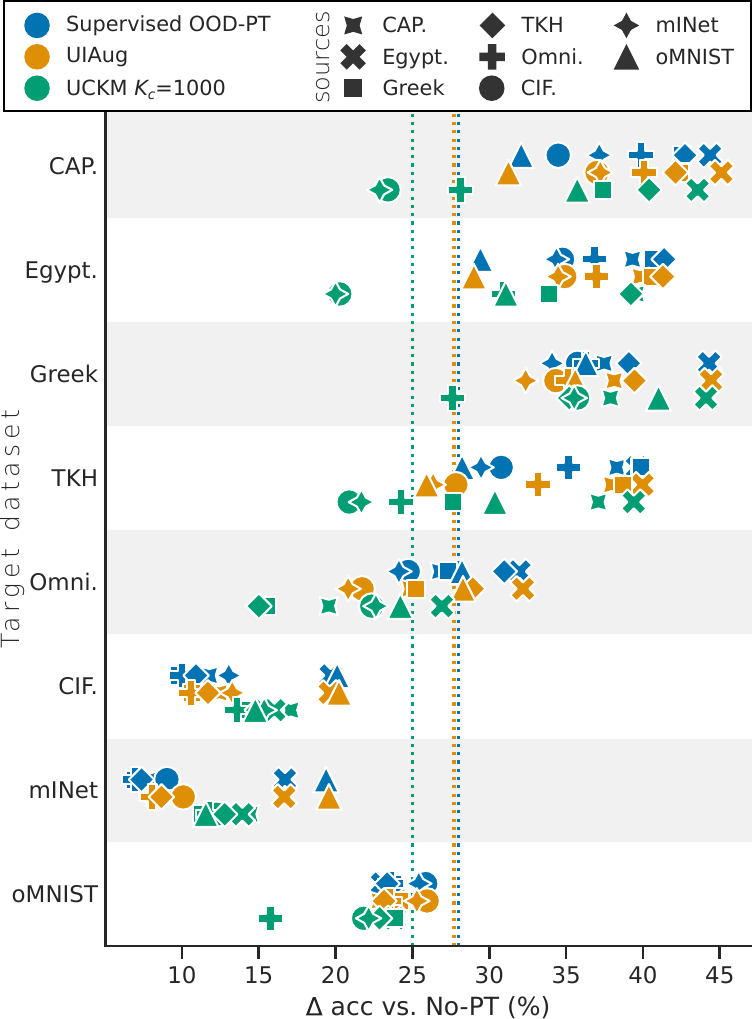}
    \caption{Accuracy gain over No-PT, $\Delta_{\mathrm{acc}}$, for each target dataset under supervised OOD-PT, \ac{UIAug}, and \ac{UCKM} with $K_c=1000$ when \ac{IDA} is applied. Marker shapes denote the OOD source domain, colors indicate the pre-training strategy, and dotted vertical lines show the mean gain for each strategy.}
    \label{fig:results:dotsAll_withIDA}
\end{figure}

Figure~\ref{fig:results:dotsAll_withIDA} compares \ac{UIAug}, supervised OOD-PT, and \ac{UCKM} with $K_c=1000$ when \ac{IDA} is applied. Results are reported as $\Delta_{\mathrm{acc}}$ over No-PT, with dotted vertical lines indicating mean gains and individual points showing results for each OOD source domain.

Overall, \ac{UIAug} closely matches supervised OOD-PT, with mean gains of 27.71 and 27.97 pp., respectively, across all source--target pairs and experimental configurations. This indicates that instance-based unsupervised pre-training recovers nearly all the benefit of using source labels. By comparison, \ac{UCKM} achieves a lower mean gain of 24.98 pp., suggesting that clustering-based pseudo-labels provide less consistent supervision in this setting.

The same conclusion holds when selecting the best OOD source for each target. \ac{UIAug} achieves an average gain of 33.59 pp., slightly above supervised OOD-PT at 33.40 pp. and above \ac{UCKM} at 31.12 pp. This further supports \ac{UIAug} as an effective label-free OOD pre-training strategy. However, the gap between the average and best-source performance confirms the strong sensitivity of OOD-PT to source--target compatibility and motivates the need for source-domain selection strategies.

The results also follow the trends observed above. Domain-specific targets show the largest gains, confirming that they benefit most from external pre-training. \ac{UIAug} closely matches supervised OOD-PT across targets, reproducing both its gain patterns and the ordering of source domains from lowest to highest gain. This correlation is weaker for \ac{UCKM}, whose source ranking varies more and whose best source--target pairs remain below those of \ac{UIAug} and supervised OOD-PT. These findings further highlight the role of source--target compatibility and confirm \ac{UIAug} as a competitive label-free alternative. Results without \ac{IDA}, reported in App.~\ref{app:UIAug_without_IDA}, show the same overall trends.

\subsection{Source-domain selection for OOD-PT}

The previous analysis in Fig.~\ref{fig:results:dotsAll_withIDA} shows that OOD-PT effectiveness strongly depends on the pre-training source domain. Across the three OOD-PT strategies, selecting the best source yields an average improvement of 11.2 pp. over selecting the worst source, with this gap reaching 20.7 pp. in the most sensitive setting. This motivates this section, which examines whether the suitability of a candidate $P_{src}$ for a target $P_{tgt}$ can be estimated before pre-training. To address this question, we evaluate similarity descriptors grouped into four families: dataset-level properties, representation-space overlap, distance-based compactness, and reconstruction-based mismatch. Their definitions are provided in Sec.~\ref{sec:method:source_selection} and App.~\ref{app:similarity-metrics}.

To isolate the effect of $P_{\mathrm{src}}$, results are averaged over the three \ac{OOD}-PT strategies and computed without \ac{IDA}, preventing target-domain adaptation from compensating for or masking the effect of source-domain choice.

\subsubsection{Individual-descriptor correlations}

\begin{figure}[ht]
    \centering
    \includegraphics[width=0.7\columnwidth]{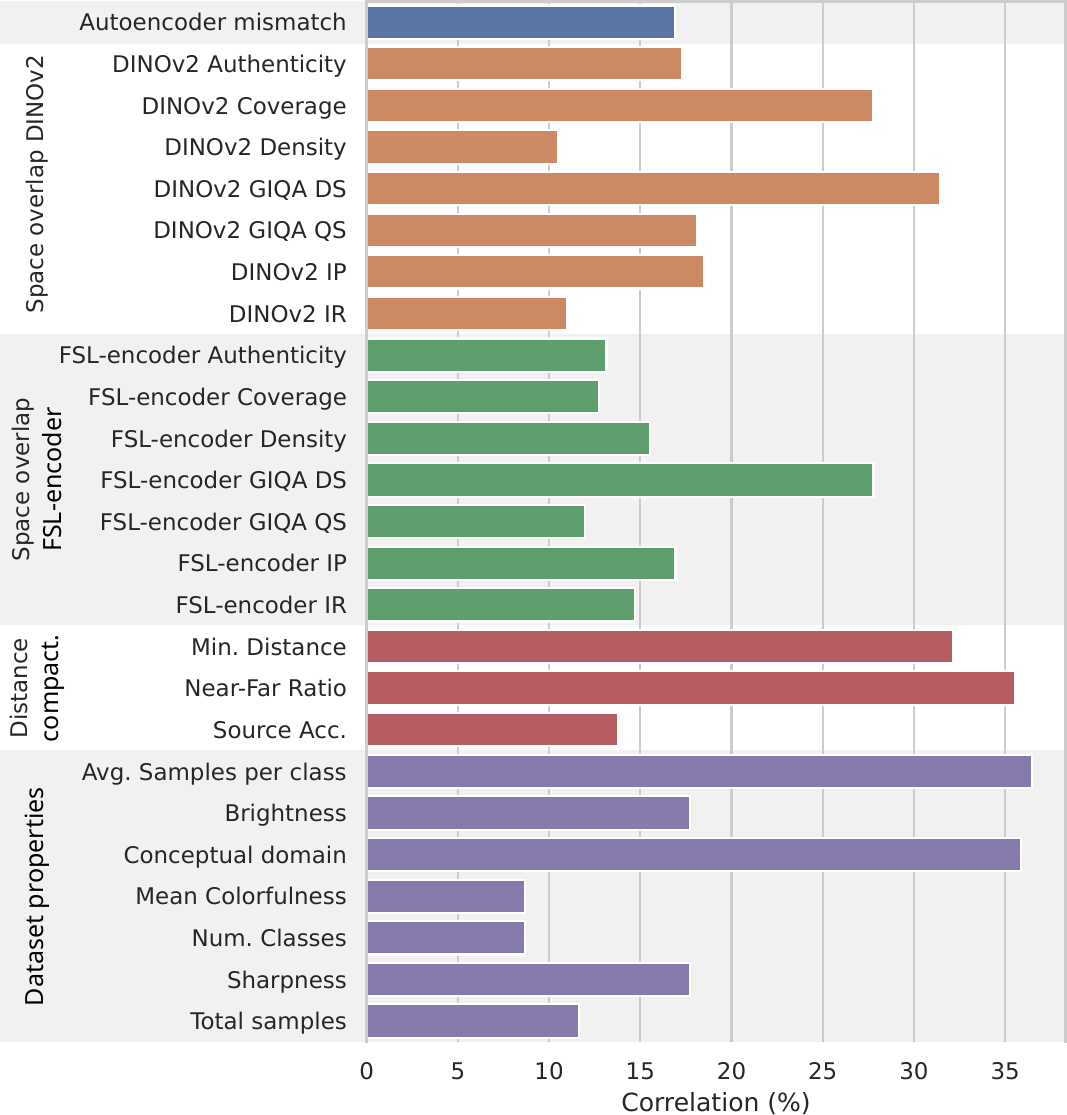}
    \caption{Pearson correlation (\%) between each descriptor family and the downstream accuracy gain obtained after OOD-PT. Correlations are computed over source--target pairs using the proposed descriptors and the observed gain with respect to the corresponding No-PT baseline.}
    \label{fig:results:correlationGroupedDelta}
\end{figure}

We first assess each descriptor independently to determine whether it is sufficiently informative for source-domain selection without requiring pre-training. Figure~\ref{fig:results:correlationGroupedDelta} reports the Pearson correlation between each descriptor family and the downstream accuracy gain after OOD-PT.

Overall, individual descriptors show only moderate predictive power. Dataset-level properties perform best, with the average number of samples per class and the conceptual-domain indicator reaching 36.5\% and 35.9\%, respectively. Distance-based compactness shows similar results, with near-far and minimum-distance correlations of 35.5\% and 32.1\%. Among representation overlap descriptors, DINO-based GIQA diversity performs best at 31.4\%, whereas FSL-encoder descriptors remain below 28\%. Autoencoder reconstruction loss reaches only 16.9\%.

These results indicate that several individual descriptors appear to reflect aspects of source-domain suitability, but none fully characterizes the source--target compatibility underlying transfer performance. We therefore examine whether factor analysis can combine these complementary descriptors into a more informative score.

\subsubsection{Factor-analysis composite score}

\begin{figure}[ht]
\centering
\includegraphics[width=0.4\columnwidth]{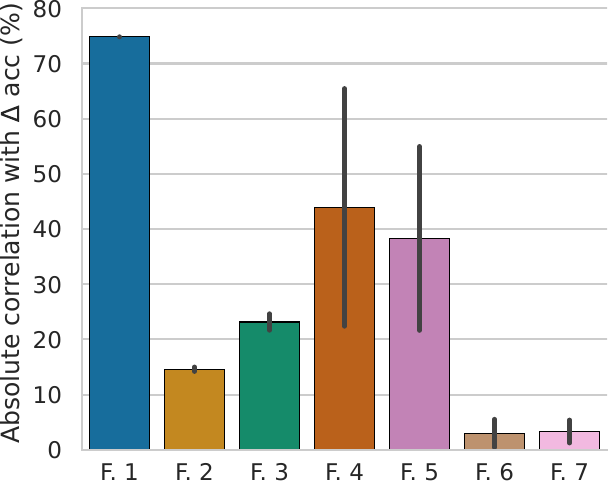}
\caption{Pearson correlation (\%) of each extracted factor with the downstream accuracy gain after OOD-PT.}
\label{fig:results:all_factors}
\end{figure}

This section evaluates the factor-analysis score described in Sec.~\ref{sec:method:factor_analysis}, which combines the descriptors into latent factors and selects the one most associated with transfer gains. Figure~\ref{fig:results:all_factors} reports the correlation between the seven extracted factors and the observed accuracy gains. The first factor is the most predictive, with a Pearson correlation of 74.8\%, compared with 43.9\% and 38.3\% for the next two. The remaining factors show much weaker associations. We therefore use the first factor as the \textit{Best factor} in the subsequent source-selection analysis.

\begin{figure}[ht]
\centering
\includegraphics[width=0.6\columnwidth]{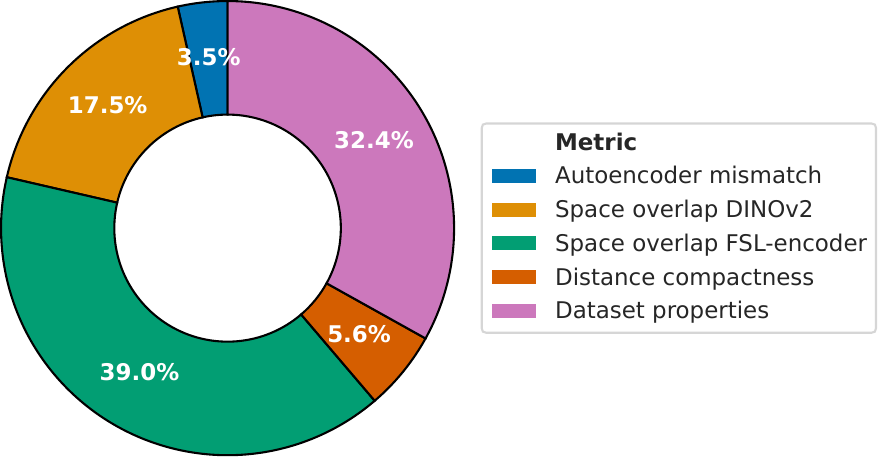}
\caption{Relative contribution of each descriptor family to the selected \textit{Best factor}. Contributions are obtained by aggregating the factor loadings of the descriptors belonging to each family and normalizing them across groups.}
\label{fig:results:loadings}
\end{figure}

To better understand what information is captured by the \textit{Best factor}, Fig.~\ref{fig:results:loadings} reports the contribution of each descriptor family. Representation-space overlap descriptors dominate, especially those based on the FSL encoder, which contribute nearly 40\% of the total. Dataset-level properties also contribute substantially, showing that source-domain suitability depends on both feature-space similarity and coarse dataset characteristics. Distance-based compactness and reconstruction-based mismatch contribute only marginally.

\subsubsection{Final analysis of source selection}

\begin{figure}[ht]
\centering
\includegraphics[width=0.7\columnwidth]{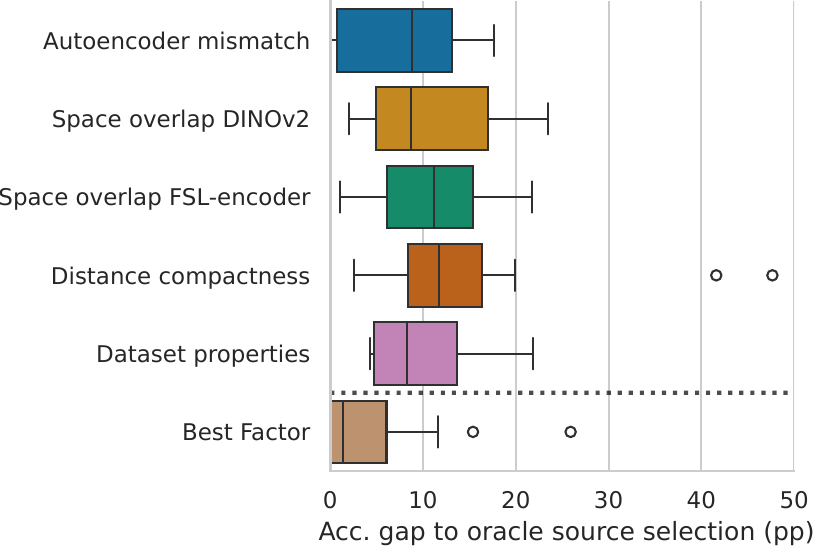}
\caption{Evaluation of source-domain selection. Candidate sources are ranked using each descriptor family or factor-based score, and the top-ranked source is selected for OOD-PT. Box plots show the downstream accuracy-gain gap in pp. between the selected and oracle sources, defined as the best-performing source after evaluating all candidates. Lower values indicate better selection.}
\label{fig:results:simulation}
\end{figure}

Finally, we evaluate source selection using the proposed descriptors. For each target, each descriptor family and the \textit{Best factor} rank the candidate sources and select the top-ranked one. We then compare the downstream accuracy gain $\Delta_\mathrm{Acc}$ obtained with this source against that of the oracle source, defined as the candidate yielding the highest gain after pre-training and target evaluation. Figure~\ref{fig:results:simulation} summarizes the resulting oracle gaps across all targets using one box plot per descriptor family and for the \textit{Best factor}.

The results show that the \textit{Best factor} provides a more reliable selection criterion than any individual descriptor family, with a median oracle gap of only 1.37 pp. All individual families yield median gaps above 8 pp. A descriptor-level breakdown is provided in App.~\ref{sec:appendix:source-selection}; although some individual descriptors achieve competitive results, the analysis confirms the same overall trend. Thus, no single descriptor captures source--target suitability sufficiently well, whereas combining complementary signals is key to reducing the error of automatic source selection.

\begin{figure}[ht]
\centering
\includegraphics[width=0.9\columnwidth]{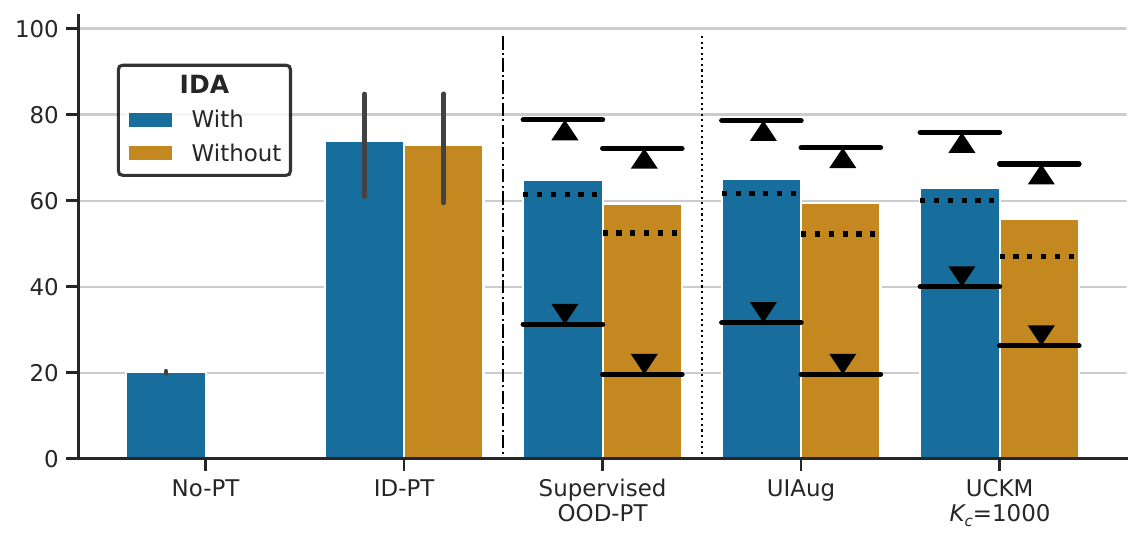}
\caption{Comparison of pre-training regimes. Bars report mean few-shot accuracy for No-PT, ID-PT, and the three OOD-PT strategies. For OOD-PT, bars show the performance obtained with the source selected by the \textit{Best factor}, dashed lines indicate the average over all candidate sources, while upper and lower arrows mark the best and worst choices, respectively.}
\label{fig:general_experiments:simpler}
\end{figure}

To contextualize this selection behavior in terms of final few-shot performance, Fig.~\ref{fig:general_experiments:simpler} reports the aggregate accuracy across pre-training regimes. For the three OOD-PT strategies, the bars show the performance obtained using the source selected by the \textit{Best factor}, the dashed lines show the average over all candidate sources, and the upper and lower arrows indicate the performance obtained with the best and worst sources, respectively.

The automatic selection based on the \textit{Best factor} consistently outperforms the average OOD source choice. Without \ac{IDA}, it improves average accuracy by 6.8--8.7 pp. across OOD-PT strategies and exceeds the worst source choice by 29.5--39.8 pp. With IDA, these gains decrease to 3.0--3.5 pp. over the average and 23.0--33.6 pp. over the worst source, as target-domain adaptation partly compensates for source--target mismatch. Nevertheless, the selected source remains above the average in all settings, showing that descriptor-based selection is beneficial and avoids the substantial degradation caused by poor source choices.

The figure also confirms the main experimental trends. ID-PT achieves the highest overall accuracy, reflecting the optimistic advantage of target-domain pre-training. \ac{UCKM} performs below supervised OOD-PT, particularly without \ac{IDA}, but exhibits higher lower bounds and narrower source-dependent ranges. This suggests reduced sensitivity to the choice of source dataset, albeit at the cost of a lower performance ceiling. \ac{UIAug} provides the strongest label-free results, closely matching supervised OOD-PT both with and without IDA. Overall, the large gap between the upper and lower arrows confirms the importance of source choice for OOD-PT.

\section{Conclusions and Future Work}
\label{sec:conclusions}

This paper revisits the common assumption in standard \ac{FSL} evaluation that class-disjoint pre-training is sufficient to define a low-data learning problem. Our results show that this assumption is incomplete. Although \ac{ID} pre-training avoids label overlap with the target classes, it still exposes the model to the same visual domain as the test episodes, producing a substantial optimistic bias. In our experiments, ID-PT improves over No-PT by 33.41 pp. on average, compared with 23.75 pp. for supervised OOD-PT, revealing an average 9.66 pp. advantage associated with ID-PT. This gap shows that pre-training is not a secondary implementation detail, but a major determinant of few-shot accuracy.

In realistic \ac{FSL} scenarios, access to a large class-disjoint auxiliary dataset matching the target tasks in domain and distribution cannot be assumed, since data scarcity often characterizes the target domain itself. OOD pre-training is therefore more practical given the availability of numerous large external datasets. Our results show that OOD-PT generally outperforms No-PT, although its effectiveness depends on source--target compatibility. \ac{IDA} partially mitigates this mismatch, particularly with more support samples, but does not eliminate domain shift, indicating that realistic \ac{FSL} evaluation should consider both class separation and source--target domain compatibility.

We also show that labeled auxiliary data are not strictly necessary for effective OOD pre-training. \ac{UIAug} nearly matches supervised OOD-PT, achieving an average gain of 27.71 pp. compared with 27.97 pp., and slightly surpassing it in some cases. This makes label-free pre-training a practical alternative when source annotations are unavailable, while also expanding the pool of datasets that can be used for pre-training.

Finally, our descriptor-based source-selection strategy provides a reliable way to estimate the suitability of a source domain before pre-training, reducing the risk of negative transfer. The \textit{Best factor} remains above the average source in all settings and achieves a median gap of only 1.37 pp. to the oracle selection, while outperforming the worst source by up to 39.8 pp. without \ac{IDA} and 33.6 pp. with \ac{IDA}. Thus, effective source selection can be performed without training on every candidate dataset, while avoiding the substantial degradation caused by poor source choices.

Future work should extend this analysis to larger and more heterogeneous benchmarks, more robust unsupervised pre-training methods, and evaluation protocols that explicitly control domain overlap between pre-training and target episodes.









\section*{Acknowledgments}
This work was supported by the Generalitat Valenciana (GV), Conselleria d'Educació, Cultura, Universitats i Ocupació, through the SmallOMR project (CIAICO/2023/255). Alejandro Galan-Cuenca acknowledges the support from Programa I+D+i de la GV through grant CIACIF/2023/090.

\bibliographystyle{IEEEtran}
\bibliography{references}

%



\clearpage
\appendix


\newpage

\renewcommand{\thefigure}{\Alph{section}.\arabic{figure}}
\renewcommand{\thetable}{\Alph{section}.\arabic{table}}

 

\section{Similarity metrics}
\label{app:similarity-metrics}

This appendix details the domain-similarity descriptors used in Sec.~\ref{sec:method:source_selection} to rank candidate \ac{OOD} source domains before pre-training. Given a target domain $P_{\mathrm{tgt}}$ and a candidate source domain $P_{\mathrm{src}}^{(m)}$, each proposed descriptor captures a different source--target property that can be computed without training a source-specific model. For each source--target pair, the descriptors defined below are collected into the vector 
$\mathbf{s}_{m,\mathrm{tgt}} = [s^{(1)}_{m,\mathrm{tgt}},\ldots,s^{(J)}_{m,\mathrm{tgt}}]$, 
where $J$ denotes the total number of proposed descriptors. This vector is used by both the individual-descriptor scores and the factor-analysis score described in Sec.~\ref{sec:method:factor_analysis}. 
Unless otherwise stated, all descriptors are standardized; descriptors that naturally measure dissimilarity are sign-aligned so that larger values consistently indicate higher expected source--target compatibility.

We organize the descriptors into four families: dataset-level statistics, representation-space overlap, distance-based compactness, and reconstruction-based mismatch.

\subsection{Dataset-level statistics}

The first family captures global properties of each dataset. Each domain is characterized by three low-level image statistics---brightness, sharpness, and colorfulness---and four dataset-level metadata descriptors: the number of classes, the total number of samples, the average number of samples per class, and the high-level semantic domain category defined in Table~\ref{tab:datasets}: music notation, historical symbols, handwritten characters, medical imaging, or natural images.

Let $\eta^{(j)}(P)$ denote the value of the $j$-th property of domain $P$. For numerical properties, the corresponding source--target descriptor is defined as the absolute difference
\begin{equation}
s^{(j)}_{m,\mathrm{tgt}}
=
\left|
\eta^{(j)}\!(P_{\mathrm{src}}^{(m)})
-
\eta^{(j)}\!(P_{\mathrm{tgt}})
\right|.
\label{eq:dataset_stat_distance}
\end{equation}
For the semantic domain category, we instead use a binary mismatch indicator, which is equal to 0 when the source and target belong to the same category and 1 otherwise.

These descriptors are simple to compute and capture global visual and dataset-structure differences. However, they do not measure the overlap or separability of source and target samples in a feature space, which is addressed by the representation-based descriptors below.

\subsection{Representation-space overlap}

The second family compares source and target domains in a feature space induced by a fixed encoder $\phi(\cdot)$. Given embedded samples from $P_{\mathrm{src}}^{(m)}$ and $P_{\mathrm{tgt}}$, we compute a set of feature-based descriptors originally introduced for auditing real and synthetic datasets~\cite{faccoco:SoftwareX:2025}. In this work, we use them as source--target compatibility measures by treating the target domain as the reference distribution and the candidate source domain as the evaluated distribution.

Specifically, we consider Authenticity, Coverage, Density, Generated Image Quality Assessment Diversity Score (GIQA-DS), GIQA quality score (GIQA-QS), Improved Precision (IP), and Improved Recall (IR). These descriptors address related notions of fidelity and coverage but differ in their geometric estimators and aggregation rules. IP measures the fraction of source samples lying within the estimated target manifold, whereas Density quantifies how many target neighborhoods contain each source sample. GIQA-QS provides a complementary parametric fidelity estimate by evaluating source samples under a Gaussian mixture model fitted to the target embeddings. Conversely, IR and Coverage quantify how well the source spans the target manifold using neighborhood-based criteria, while GIQA-DS estimates this reciprocal relation by evaluating target samples under a density model fitted to the source embeddings. Finally, Authenticity measures whether source samples are close to individual target samples relative to the local spacing within the target set, thereby capturing local duplication rather than general source--target proximity. Together, these descriptors characterize complementary aspects of representation-space compatibility.

In our experiments, these descriptors are computed in two types of feature spaces: (i) a generic self-supervised representation obtained with DINOv2~\cite{Oquab:arxiv:2023}, following the original publication, and (ii) task-adapted representations extracted from the few-shot architectures considered in this work (see Sec.~\ref{sec:setup}). In the latter case, a single model is trained using only the $n$-way $k$-shot support set of the target domain; this model is then kept fixed and used as a common encoder to extract representations for both the target and candidate source domains. The DINOv2 features provide a general-purpose estimate of representation-space overlap, whereas the few-shot encoders provide a task-adapted estimate conditioned on the target support set.

\subsection{Distance-based compactness}

The third family characterizes how compact and separable each candidate source domain appears under a target-adapted few-shot representation. These descriptors do not capture an intrinsic property of the source, but rather how its samples are organized in the representation induced by the target distribution.

For a given target domain $P_{\mathrm{tgt}}$, we train a few-shot model with encoder $f_{\theta_{\mathrm{tgt}}}$ on its $n$-way $k$-shot support set and keep it fixed to extract embeddings from all candidate source domains. Given a source episode $T=(S_T,Q_T)$ sampled from a candidate source domain $P_{\mathrm{src}}^{(m)}$, let $\mathcal{C}_T$ denote its set of classes or pseudo-classes. For supervised \ac{OOD}-PT, $\mathcal{C}_T$ is defined by the ground-truth source labels, whereas for the label-free variants it is defined by the pseudo-labels induced by the corresponding strategy (see Sec.~\ref{sec:method:labelfree}). Following the general predictor in Eq.~\eqref{eq:fsl_predictor}, each source query is evaluated through the target-adapted model as
\begin{equation}
h_{\theta_{\mathrm{tgt}},T}(x)
=
g_T\big(f_{\theta_{\mathrm{tgt}}}(x); S_T\big),
\qquad x\in Q_T ,
\end{equation}
where $g_T$ is the architecture-specific episodic classifier induced by the source support set $S_T$. During this inference process, for each source query image $x \in Q_T$, we collect the architecture-specific distances or dissimilarity scores computed between its embedding and the support-derived embeddings or class representatives used by $g_T$, denoted by $\mathcal{D}_{\theta_{\mathrm{tgt}}}(x,S_T)$. From these scores, we derive three complementary descriptors.

The first descriptor measures the contrast between near and far candidate classes. Let $q_{50}$ be the empirical median of $\mathcal{D}_{\theta_{\mathrm{tgt}}}(x,S_T)$. We split the scores into a lower half, containing values up to the median, and an upper half, containing values above the median, and define the query-level \textit{near-to-far distance ratio} as
\begin{equation}
\rho_{\mathrm{Q}}(x,S_T)
=
\frac{
\sum_{d_i\in \mathcal{D}_{\theta_{\mathrm{tgt}}}(x,S_T):\, d_i \le q_{50}} d_i
}{
\sum_{d_i\in \mathcal{D}_{\theta_{\mathrm{tgt}}}(x,S_T):\, d_i > q_{50}} d_i
+ \epsilon
}.
\label{eq:quartile_distance_ratio}
\end{equation}
where $\epsilon>0$ is a small constant for numerical stability. Lower values indicate a more contrasted distance profile, in which the query is much closer to a subset of candidate classes than to the remaining ones, whereas higher values correspond to a flatter profile and therefore suggest weaker separability under the target-adapted representation. The final descriptor is obtained by averaging $\rho_{\mathrm{Q}}(x,S_T)$ over query samples and sampled source episodes.

The second descriptor keeps only the \textit{minimum distance} for each query, i.e., the distance to its nearest source representative, and aggregates this value over queries and episodes. This provides a direct nearest-prototype compactness measure, where lower values indicate that source samples tend to lie close to at least one source class representative under the target-adapted embedding.

Finally, we compute the \textit{source accuracy}, defined as the episodic classification accuracy obtained when the target-trained few-shot model classifies source-domain query samples. Accuracy is measured against the available source labels in the supervised case or against pseudo-labels in the label-free case. This descriptor complements the geometric distance summaries by assessing whether target-adapted compactness translates into correct source-domain classification.

\subsection{Reconstruction-based mismatch}

The fourth family estimates source--target visual mismatch from reconstruction error, treating the comparison as an anomaly-detection problem. Following reconstruction-based anomaly-detection methods~\cite{Yang2022anomaly, Bergmann2019anomaly}, we assume that an autoencoder fitted to the target support set should reconstruct samples from visually similar source domains with lower error. 
Thus, for a target-domain setting, we train a single convolutional autoencoder $a_{\psi}^{\mathrm{tgt}}$ to reconstruct the images in the $n$-way $k$-shot target support set, without using their labels. This autoencoder is then kept fixed and used to evaluate all candidate source domains. For each source domain $P_{\mathrm{src}}^{(m)}$, we compute the reconstruction-based mismatch descriptor as
\begin{equation}
D_{\mathrm{AE}}\!\left(P_{\mathrm{src}}^{(m)},P_{\mathrm{tgt}}\right)
=
\frac{1}{|\mathcal{X}_{\mathrm{src}}^{(m)}|}
\sum_{x_i\in\mathcal{X}_{\mathrm{src}}^{(m)}}
\frac{1}{HWC}
\left\|
a_{\psi}^{\mathrm{tgt}}(x_i)-x_i
\right\|_2^2 ,
\label{eq:ae_visual_divergence}
\end{equation}
where $\mathcal{X}_{\mathrm{src}}^{(m)}$ denotes the images from the $m$-th candidate source domain, and $H$, $W$, and $C$ are the image height, width, and number of channels. Low values indicate that source samples are compatible with the reconstruction patterns learned from the target support set, suggesting a low visual mismatch.

In our implementation, the autoencoder operates on grayscale input images. The encoder consists of four $3\times3$ convolutional layers with channel sizes $16$, $32$, $64$, and $128$, each followed by a LeakyReLU activation and $2\times2$ max-pooling. The resulting feature map is flattened and projected to a 512-dimensional latent representation. The decoder maps this latent code back to a feature map and reconstructs the image through four transposed-convolution layers with channel sizes $64$, $32$, $16$, and $1$, using LeakyReLU activations except for the final sigmoid output. The model is trained with a mean squared reconstruction loss using Adam.

\section{Robustness of \ac{UIAug} without \ac{IDA}}
\label{app:UIAug_without_IDA}

\begin{figure}[ht]
    \centering
    \includegraphics[width=0.6\columnwidth]{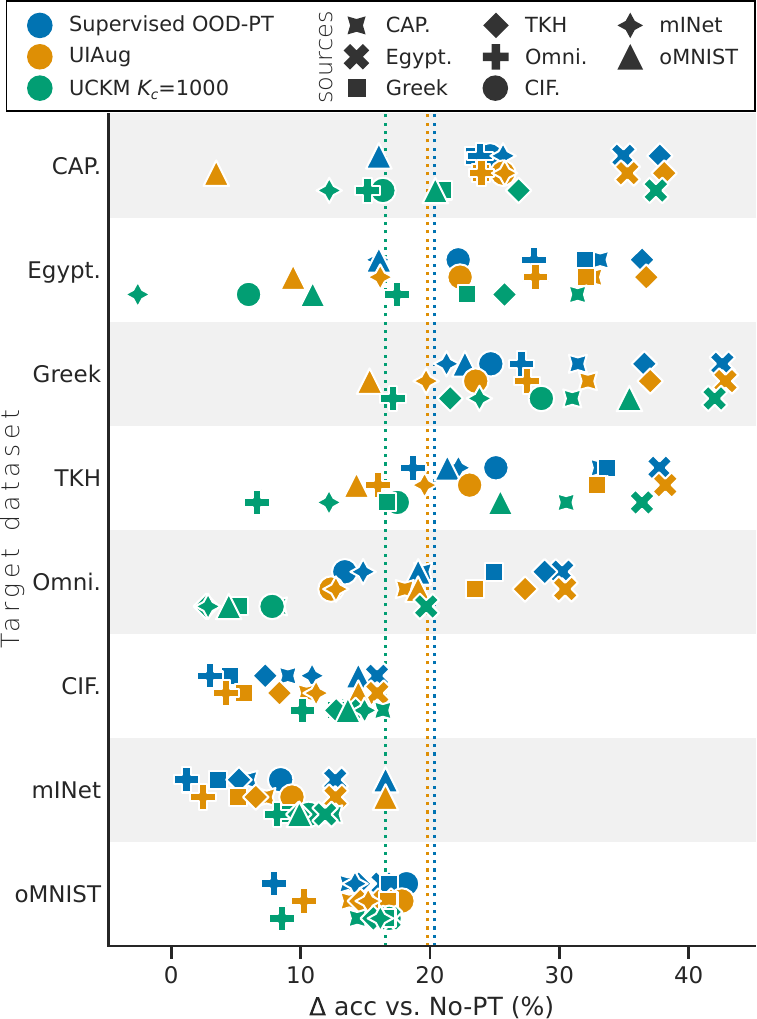}
    \caption{Accuracy gain over the corresponding No-PT baseline, $\Delta_{\mathrm{acc}}$, for supervised OOD-PT, \ac{UIAug}, and \ac{UCKM} with $K_c=1000$ without applying \ac{IDA}. Results are shown for each target dataset, where marker shapes indicate the OOD source domain and colors indicate the pre-training strategy. Dotted vertical lines denote the average gain for each strategy.}
    \label{fig:results:dotsAll_withoutIDA}
\end{figure}

Figure~\ref{fig:results:dotsAll_withoutIDA} reports the same comparison as in Figure~\ref{fig:results:dotsAll_withIDA}, but without applying \ac{IDA}. As expected, removing \ac{IDA} reduces the absolute gains with respect to the main analysis. However, the relative behavior of the three pre-training strategies is largely preserved. Averaged over all source--target combinations and experimental configurations, supervised OOD-PT obtains a mean gain of 20.35 percentage points, while \ac{UIAug} reaches 19.78 percentage points. This confirms that \ac{UIAug} remains close to its supervised counterpart even without adaptation. In contrast, \ac{UCKM} obtains a lower average gain of 16.51 percentage points.

The same conclusion holds when considering the best OOD source for each target. In this case, \ac{UIAug} reaches an average gain of 45.78 percentage points, slightly above supervised OOD-PT, which obtains 45.02 percentage points, and clearly above \ac{UCKM}, which obtains 40.04 percentage points. These results support the conclusions drawn in the main paper: \ac{UIAug} provides a competitive label-free alternative to supervised OOD pre-training, while \ac{UCKM} is less consistent. Moreover, the large gap between average and best-source performance confirms the importance of source--target compatibility, independently of the use of \ac{IDA}.

\section{Descriptor-level analysis of source selection}
\label{sec:appendix:source-selection}

\begin{figure}[ht]
    \centering
    \includegraphics[width=0.7\columnwidth]{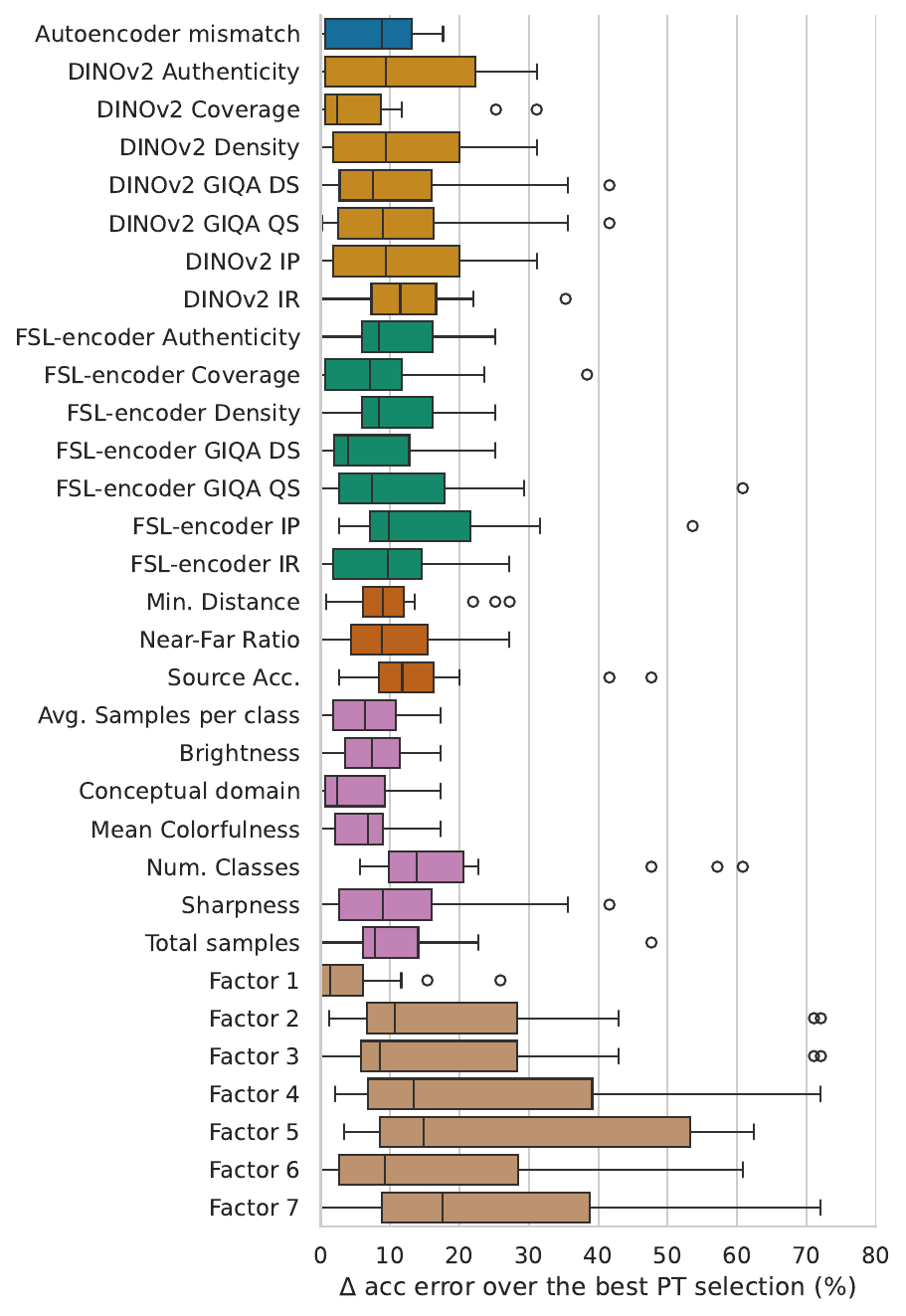}
    \caption{Descriptor-level evaluation of source-domain selection. Candidate sources are ranked using each individual descriptor or extracted factor, and the top-ranked source is selected for OOD-PT. Box plots report the downstream accuracy-gain gap in pp. between the selected source and the oracle source, defined as the best-performing candidate after evaluating all sources. Factor~1 corresponds to the \textit{Best factor} used in the main analysis. Results are averaged over the three OOD-PT strategies without \ac{IDA}. Lower values indicate better selection.}
    \label{fig:results:simulation:All}
\end{figure}

Figure~\ref{fig:results:simulation:All} provides the descriptor-level counterpart to Fig.~\ref{fig:results:simulation}. Rather than aggregating the descriptors into conceptual families, it reports the source-selection error obtained with each individual descriptor and each extracted factor. Factor~1, denoted as the \textit{Best factor} in the main analysis, achieves the lowest overall error, with a median oracle gap of 1.37 pp.

Among the individual descriptors, \textit{Conceptual domain} and \textit{DINOv2 Coverage} obtain the lowest median gaps, both at 2.41 pp., followed by \textit{FSL-encoder GIQA-DS} at 4.02 pp. The conceptual-domain indicator is also comparatively robust: by prioritizing sources from the same high-level semantic category as the target whenever available, it avoids many large errors associated with clearly mismatched source--target pairs. The competitive performance of the two representation-based descriptors further suggests that the extent to which a source covers the target representation space is informative of its transfer suitability. Nevertheless, their higher variability across targets supports the main conclusion: individual descriptors can provide effective selection cues, but combining complementary signals through factor analysis yields more reliable source selection overall.

\end{document}